\documentclass{article}

\PassOptionsToPackage{numbers, compress}{natbib}

\usepackage[preprint]{neurips_2026}

\usepackage[utf8]{inputenc} 
\usepackage[T1]{fontenc}    
\usepackage{hyperref}       
\usepackage{url}            
\usepackage{booktabs}       
\usepackage{amsfonts}       
\usepackage{nicefrac}       
\usepackage{microtype}      
\usepackage{xcolor}         
\usepackage[table,xcdraw]{xcolor}
\usepackage{graphicx}
\usepackage{array}
\usepackage{tgheros} 
\usepackage{amsmath} 
\usepackage{multirow}
\usepackage[ruled,vlined]{algorithm2e}
\usepackage{float}
\usepackage{wrapfig}

\title{ACPruner: Visual Token Pruning as Biased Attention Coverage Maximization in LVLMs}

\author{
\textbf{Xu Li} \quad
\textbf{Yuxuan Liang} \quad
\textbf{Yi Zheng} \quad
\textbf{Zhe Liu} \quad
\textbf{Xiaolei Chen} \\
\textbf{Haotian Chen} \quad
\textbf{Rui Zhu} \quad
\textbf{Fan Shi} \quad
\textbf{Xiangyang Xue} \\[2mm]
{\small\rmfamily College of Computer Science and Artificial Intelligence} \\
{\small\rmfamily Fudan University, Shanghai, China}
}

\begin{document}

\maketitle

\begin{abstract}
Large Vision-Language Models (LVLMs) face significant computational inefficiencies caused by the large number of visual tokens. Existing visual token pruning methods mainly focus on either retaining individually important tokens or selecting mutually diverse ones. In this work, we revisit visual token pruning from a coverage perspective and formulate it as a biased attention coverage maximization problem. The key idea is to select a compact token subset whose encoder-side outgoing attention can jointly cover the image while assigning higher coverage priority to more informative regions. From this perspective, we propose \textbf{ACPruner}, a training-free visual token pruning framework for efficient LVLM inference. ACPruner first estimates token importance by combining intra-modal saliency and inter-modal relevance, then derives token-wise coverage from attention patterns within the vision encoder, and finally performs greedy selection to maximize the proposed coverage objective. Extensive experiments across multiple LVLM backbones, including LLaVA-1.5-7B/13B, LLaVA-NeXT-7B/13B, Qwen2.5-VL-7B, and LLaVA-OneVision-7B, show that ACPruner consistently achieves strong performance retention while delivering substantial end-to-end inference speedups.
\end{abstract}

\section{Introduction}
By integrating vision encoders with Large Language Models (LLMs), Large Vision-Language Models (LVLMs) have achieved impressive performance across a wide range of multimodal tasks, such as visual question answering, document understanding, and general multimodal reasoning~\cite{llava1.5, llavanext, qwen2.5}. However, this strong performance often comes at the cost of processing a large number of visual tokens~\cite{mustdrop, survey1, hidrop, vica}, especially in scenarios involving high-resolution images, multi-image inputs or long videos. This token explosion substantially increases computational overhead, limiting the practicality of LVLMs in resource-constrained or latency-sensitive applications~\cite{survey2, videostreaming, videotracking}. Therefore, reducing the number of visual tokens while preserving model performance has become an increasingly important problem for efficient LVLM inference~\cite{fastv, fastervlm, visionzip, divprune, lvlm-survey}. 

Existing visual token pruning methods for LVLMs can be broadly categorized into two paradigms. \textbf{Importance-based methods} prune tokens according to token-level importance scores. Depending on how importance is defined, they can be further divided into visual saliency based~\cite{hired,vispruner,visionzip} and textual relevance based approaches~\cite{fastv,sparsevlm,pdrop}. The former typically rely on attention signals within the vision encoder to identify visually salient tokens, while the latter use crossmodal attention inside the LLM to preserve tokens most relevant to the input text. Despite using different metrics, these methods all follow a token-wise ranking paradigm that retains the highest-scoring tokens. In contrast, \textbf{diversity-based methods}~\cite{dart, divprune} focus on reducing redundancy among retained tokens. They usually compute pairwise feature similarities and retain a subset of most mutually dissimilar tokens, thereby preserving more semantically complementary visual information.

\begin{figure}[t]
    \centering
    \includegraphics[width=\columnwidth]{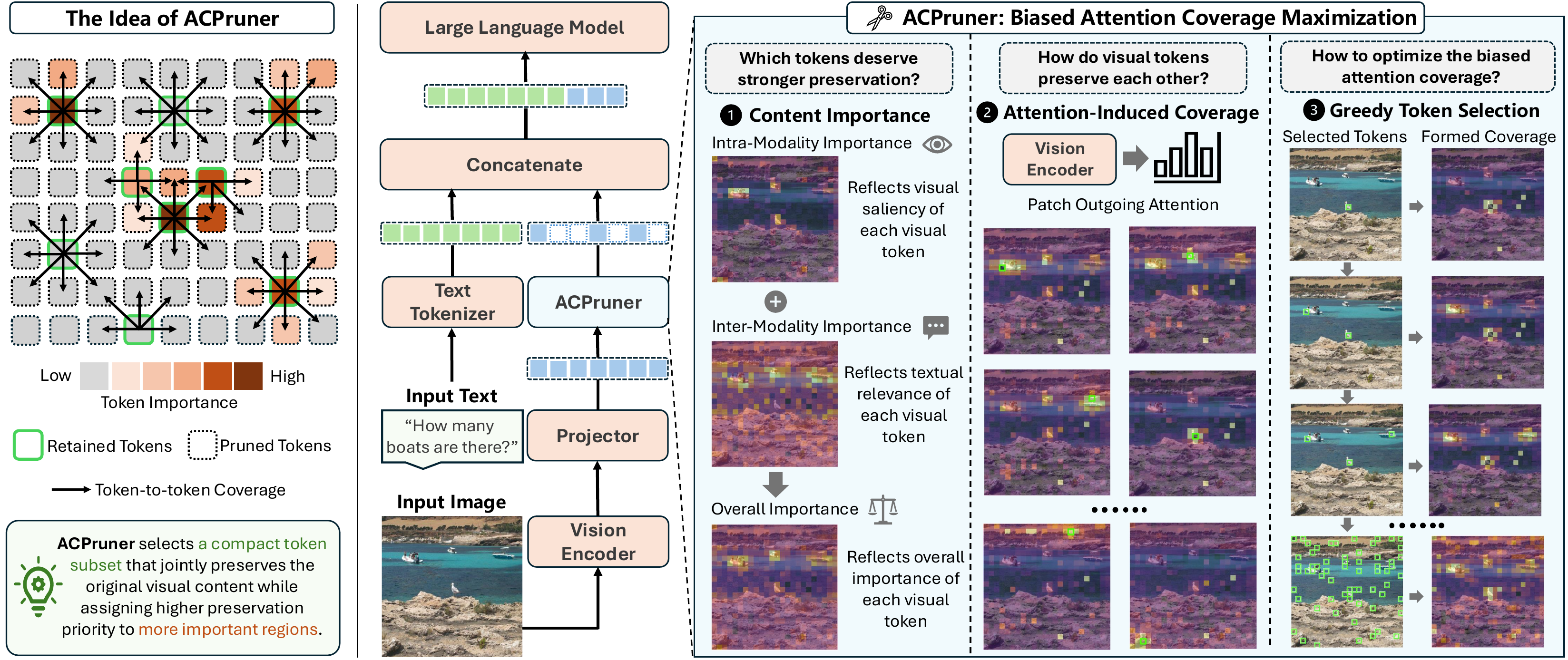}
    \caption{The left part illustrates the core idea of ACPruner. The right part shows how ACPruner is integrated with LVLM inference pipelines and provides a workflow visualization of the method.}
    \label{fig:overview}
\end{figure}

While these methods have shown empirical effectiveness, they still suffer from inherent limitations. Importance-based methods treat token selection as an individual ranking problem. As a result, the retained tokens may concentrate excessively on a few salient regions, causing redundant preservation of similar content while overlooking the surrounding context~\cite{dart,divprune}. Diversity-based methods mitigate this issue by encouraging less redundancy among selected tokens, but they do not distinguish which content is more important to preserve. Consequently, they may retain tokens that are semantically distinct yet uninformative for the downstream task~\cite{d2pruner,scope}. 

We argue that the goal of visual token pruning is neither to retain the individually most important tokens nor the most mutually dissimilar ones, but to select a compact token subset that can jointly preserve the original visual content while placing stronger emphasis on content that is more critical for downstream reasoning. This intuition can be understood through a real-world security deployment analogy. Given a limited number of guards, an effective strategy should neither concentrate all guards in a few high-risk regions without considering overall coverage, nor distribute them uniformly without considering regional importance. Instead, each guard should be placed by jointly considering both the importance of the region and the area it can cover. 

Motivated by this view, we formulate visual token pruning as a biased coverage maximization problem. The goal is to select a compact token subset that can collectively preserve the original visual information while assigning higher preservation priority to the most important content. Based on this formulation, we propose \textbf{ACPruner}, a training-free visual token pruning framework for LVLMs. As shown in Fig.~\ref{fig:overview}, ACPruner first estimates the preservation priority of each visual token by jointly modeling visual saliency and textual relevance. It then derives token-to-token coverage relationships from patch-to-patch attention within the vision encoder, capturing how one token can preserve the information of another. With these two components, ACPruner greedily selects tokens that provide the largest marginal gain in an importance-biased coverage objective. In this way, ACPruner explicitly accounts for both which visual tokens should be prioritized and how the retained tokens collectively preserve the original visual information. Moreover, since ACPruner operates before the LLM and does not rely on LLM-side attention signals, it effectively reduces the expensive LLM inference overhead while remaining compatible with FlashAttention. Extensive experiments across diverse LVLM families and task categories demonstrate that ACPruner achieves strong performance retention while delivering substantial end-to-end efficiency gains in practice. 

Overall, our contributions are threefold: (1) We reformulate visual token pruning as a biased attention coverage maximization problem, offering a new perspective beyond conventional importance-based and diversity-based formulations; (2) We propose ACPruner, a training-free and pre-LLM visual token pruning framework for LVLMs that instantiates this formulation by jointly modeling content importance and attention-induced token coverage; (3) We validate the effectiveness of ACPruner across diverse LVLM backbones, including LLaVA-1.5-7B/13B, LLaVA-NeXT-7B/13B, LLaVA-OneVision-7B and Qwen2.5-VL-7B, on both image and video understanding tasks.

\section{Method}

\subsection{Problem Definition}
A typical LVLM consists of three components: a vision encoder, a projector, and an LLM. Given an input image, the vision encoder first extracts visual features, which are subsequently mapped by the projector into the input embedding space of the LLM. We denote the resulting visual tokens as $\mathbf{V} = \{\mathbf{v}_1,\ldots,\mathbf{v}_T\} \subseteq \mathbb{R}^{d}$, where $T$ is the number of visual tokens and $d$ is the embedding dimension. Given an input textual query, we denote the embedded textual tokens as $\mathbf{U} = \{\mathbf{u}_1,\ldots,\mathbf{u}_N\} \subseteq \mathbb{R}^{d}$, where $N$ is the number of textual tokens. The visual and textual tokens are concatenated in sequence and jointly processed by the LLM to produce a textual response, denoted by $\mathcal{M}_{\theta}(\mathbf{V}, \mathbf{U})$.
The goal of visual token pruning is to select a visual token subset $\mathbf{S} \subset \mathbf{V}$, with $|\mathbf{S}| = k$ and $k \ll T$, that can replace the full visual token set while preserving the original model behavior as much as possible. This objective can be formally written as a constrained optimization problem:
\begin{equation}
\min_{\mathbf{S} \subset \mathbf{V}}
\ \mathcal{L}\!\left(\mathcal{M}_{\theta}(\mathbf{S},\mathbf{U}),\ \mathcal{M}_{\theta}(\mathbf{V},\mathbf{U})\right)
\quad \text{s.t.} \quad |\mathbf{S}| = k,
\label{eq:ideal_obj}
\end{equation}
where $\mathcal{L}$ measures the discrepancy between the model outputs before and after visual token pruning.

\subsection{Reformulating Visual Token Pruning as Biased Coverage Maximization}

Directly optimizing Eq.~\eqref{eq:ideal_obj} in a training-free setting is intractable, as it requires evaluating the effect of every candidate subset on the final model output. Therefore, a surrogate objective is needed to assess the quality of $\mathbf{S}$ at an early stage of the inference pipeline.
Existing methods usually rely on importance-based or diversity-based objectives. Importance-based methods aim to maximize the sum of individual token importance scores within $\mathbf{S}$, while diversity-based methods seek to minimize the mutual feature similarity within $\mathbf{S}$. However, both paradigms focus on the selected subset $\mathbf{S}$ itself, without explicitly modeling how well it preserves the original token set $\mathbf{V}$.

In this work, we treat each original visual token $\mathbf{v}_j \in \mathbf{V}$ as an information unit to be preserved, and ask whether the selected subset $\mathbf{S}$ can jointly preserve the overall information carried by the tokens in $\mathbf{V}$. Moreover, since different visual tokens contribute unequally to downstream reasoning, more informative content should receive higher preservation priority. Based on this view, we reformulate visual token pruning as a biased coverage maximization problem:
\begin{equation}
\max_{\mathbf{S} \subset \mathbf{V}} \ \sum_{j=1}^{T} w_j \, \mathrm{Cov}(\mathbf{S}, \mathbf{v}_j)
\quad \text{s.t.} \quad |\mathbf{S}| = k,
\label{eq:highlevel_coverage_obj}
\end{equation}
where $w_j \ge 0$ denotes the content importance of $\mathbf{v}_j$, and $\mathrm{Cov}(\mathbf{S}, \mathbf{v}_j)$ measures how well the selected subset $\mathbf{S}$ preserves the content of $\mathbf{v}_j$. Under this formulation, an effective token subset should not only preserve the original visual content collectively, but also allocate stronger preservation capacity to the most important tokens. In the following, we instantiate this objective by defining the importance weight $w_j$ and the coverage term $\mathrm{Cov}(\mathbf{S}, \mathbf{v}_j)$.

\subsection{Content Importance: Which Tokens Deserve Stronger Preservation?}

In Eq.~\eqref{eq:highlevel_coverage_obj}, the importance weight $w_j$ determines the preservation priority of each original visual token $\mathbf{v}_j \in \mathbf{V}$. We estimate this priority from two complementary perspectives: intra-modality importance, which captures how much a token contributes to the global visual semantics, and inter-modality importance, which measures how relevant the token is to the input textual query.

\paragraph{Intra-modality importance.}
To estimate the visual saliency of each token, we leverage the CLS-to-patch attention in the penultimate layer of the vision encoder, and define the intra-modality importance of token $\mathbf{v}_j$ as:
\begin{equation}
s_j^{\mathrm{intra}}
=
\frac{1}{H}\sum_{h=1}^{H} A^{(h)}_{\mathrm{cls}\rightarrow j},
\label{eq:intra_cls}
\end{equation}
where $A^{(h)}_{\mathrm{cls}\rightarrow j}$ denotes the attention weight assigned by the CLS token to the $j$-th patch token at head $h$, and $H$ is the number of attention heads. For LVLMs whose vision encoders do not contain a CLS token, we use the average incoming attention of each patch token as a surrogate saliency measure.

\paragraph{Inter-modality importance.}
To measure the textual relevance of each visual token, we first use a lightweight NLP model~\cite{en_core_web_sm} to extract nouns from the input text and treat them as the key semantic units of the query. For each noun, we obtain a word-level embedding by average pooling its token embeddings. Let $\mathcal{N}(\mathbf{U}) \subseteq \mathbb{R}^{d}$ denote the set of extracted word-level embeddings. The inter-modality importance of $\mathbf{v}_j$ is defined as:
\begin{equation}
s_j^{\mathrm{inter}}
=
\max_{\mathbf{u}_k \in \mathcal{N}(\mathbf{U})} r_{jk},
\label{eq:inter_score}
\end{equation}
where $r_{jk}$ denotes the cosine similarity between visual token $\mathbf{v}_j$ and word embedding $\mathbf{u}_k$. Since this computation does not rely on LLM-side attention weights, it remains compatible with FlashAttention.

\paragraph{Fused importance.}
We combine the two importance signals to obtain the final importance weight:
\begin{equation}
w_j
=
\alpha \, \tilde{s}_j^{\mathrm{intra}}
+
(1-\alpha)\,\tilde{s}_j^{\mathrm{inter}},
\label{eq:fused_importance}
\end{equation}
where $\tilde{s}_j^{\mathrm{intra}}$ and $\tilde{s}_j^{\mathrm{inter}}$ denote the softmax-normalized intra-modality and inter-modality importance scores, and $\alpha \in [0,1]$ controls the trade-off between them. When noun extraction returns an empty set, we fall back to pure intra-modality importance by setting $\alpha=1$.

\subsection{Attention-Induced Coverage: How Do Visual Tokens Preserve Each Other?}

LVLMs typically use ViT-based vision encoders~\cite{vit}, where visual tokens interact with each other through self-attention. Therefore, each visual token does not only encode its local patch content, but also aggregates information from other tokens~\cite{patchinteraction1}. This motivates us to use patch-to-patch attention inside the vision encoder to characterize how one visual token can help preserve another.

Let $A^{(l,h)}_{i \rightarrow j}$ denote the attention weight from token $\mathbf{v}_i$ to token $\mathbf{v}_j$ at layer $l$ and head $h$ in the vision encoder. We define the token-level coverage of $\mathbf{v}_i$ over $\mathbf{v}_j$ as:
\begin{equation}
c_{ij}
=
\frac{1}{LH}
\sum_{l=1}^{L}
\sum_{h=1}^{H}
A^{(l,h)}_{i \rightarrow j},
\label{eq:token_coverage}
\end{equation}
where $L$ and $H$ denote the numbers of layers and attention heads. It measures how much token $\mathbf{v}_i$ attends to token $\mathbf{v}_j$ on average, and thus how much $\mathbf{v}_i$ preserves the information associated with $\mathbf{v}_j$. Based on this token-level coverage relationship, we define the coverage of a selected subset $\mathbf{S}$ over an original token $\mathbf{v}_j$ as: 
\begin{equation}
\mathrm{Cov}(\mathbf{S}, \mathbf{v}_j)
=
\begin{cases}
\max_{\mathbf{v}_i \in \mathbf{S}} c_{ij}, & \text{if } \mathbf{S}\neq\emptyset,\\
0, & \text{if } \mathbf{S}=\emptyset.
\end{cases}
\label{eq:set_coverage}
\end{equation}
This attention-induced coverage differs fundamentally from relations defined by pairwise feature similarity. Feature similarity measures static proximity in the embedding space, whereas our coverage is derived from the internal information flow of the vision encoder. Consequently, the proposed coverage is generally directional, i.e., $c_{ij} \neq c_{ji}$ in general. Such asymmetry is desirable because preservation is inherently directional: one retained token may strongly preserve another, while the reverse relation can be much weaker.

\subsection{Optimizing the Biased Attention Coverage Objective}

With the importance weight and coverage relation defined above, we instantiate the final pruning objective as:
\begin{equation}
\max_{\mathbf{S} \subset \mathbf{V}}
\ \sum_{j=1}^{T} w_j^{\beta} \, \mathrm{Cov}(\mathbf{S}, \mathbf{v}_j)
\quad \text{s.t.} \quad |\mathbf{S}| = k,
\label{eq:final_objective}
\end{equation}
where $\beta > 0$ controls the strength of the importance bias. 
This objective encourages the selected subset to collectively preserve the original token set through attention-induced coverage, while allocating greater preservation priority to more important visual content.

Directly solving Eq.~\eqref{eq:final_objective} is NP-hard (proof provided in Appendix~\ref{appendix:nphard}), we therefore adopt a greedy selection strategy. Given a current selected subset $\mathbf{S}$, we define the marginal coverage gain of a candidate token $\mathbf{v}_i \in \mathbf{V} \setminus \mathbf{S}$ as:
\begin{equation}
\Delta(\mathbf{v}_i\mid \mathbf{S})
=
\sum_{j=1}^{T}
w_j^{\beta} \,
\max\!\left(c_{ij} - m_j,\, 0\right),
\label{eq:marginal_gain}
\end{equation}
where $m_j=\mathrm{Cov}(\mathbf{S},\mathbf{v}_j)$ denotes the current coverage of subset $\mathbf{S}$ over token $\mathbf{v}_j$. This marginal gain measures how much additional importance-biased coverage the candidate token $\mathbf{v}_i$ can provide for content that is not yet well preserved. At each step, we select the token with the largest marginal gain and add it to $\mathbf{S}$ until the token budget $k$ is reached. The pseudocode is provided in Appendix~\ref{appendix:pseudocode}.

In Appendix~\ref{appendix:theory}, we further show that Eq.~\eqref{eq:final_objective} is equivalent to minimizing an importance-weighted representative distortion over the original visual token set, and that the objective is monotone submodular. Therefore, the greedy algorithm admits a standard approximation guarantee, indicating that ACPruner is grounded in a principled optimization formulation rather than a heuristic design.

\section{Experiments}

\vspace{-4pt}
\begin{table}[h]
\caption{Performance comparison on LLaVA-1.5-7B under different pruning ratios. ``Perf. Ret.'' denotes the average performance retention rate relative to the full-token baseline.
}
\label{tab:llava-1.5-7b}
\resizebox{\columnwidth}{!}{
\begin{tabular}{l|c|ccccccccc|c}
\toprule
\rowcolor[HTML]{F2F3F5} 
\textbf{Method}   & \textbf{Venue} & \textbf{MME}  & \textbf{GQA}  & \textbf{SQA} & \textbf{POPE} & \textbf{TextVQA} & \textbf{VizWiz} & \textbf{VQA-v2} & \textbf{MMB} & \textbf{MM-Vet} & \textbf{Perf. Ret.} \\
\midrule
\rowcolor[HTML]{F5F6F7}
\multicolumn{12}{c}{\cellcolor[HTML]{F5F6F7}Upper Bound, 576 visual tokens  (pruning ratio = 0\%)}                                                                                                 \\
\midrule
LLaVA-1.5-7B      & CVPR'24       & 1862          & 61.9          & 69.5           & 85.9          & 58.2             & 50.0            & 78.4            & 64.7            & 31.3            & 100.0\%             \\
\midrule
\rowcolor[HTML]{F5F6F7}
\multicolumn{12}{c}{\cellcolor[HTML]{F5F6F7}Retain 192 visual tokens (pruning ratio = 66.7\%)}                                                                                                     \\
\midrule
FastV             & ECCV'24       & 1612          & 52.7          & 67.3           & 64.8          & 52.5             & 50.8            & 67.1            & 61.2            & 27.7            & 89.4\%              \\
PruMerge          & ICCV'25       & 1632          & 54.3          & 67.9           & 71.3          & 54.3             & 50.1            & 70.6            & 59.6            & -               & 91.5\%              \\
FasterVLM         & arXiv'24      & 1780          & 59.3          & 70.0           & 85.3          & 57.3             & 50.1            & 75.2            & 63.5            & 31.8            & 98.4\%              \\
VisPruner         & ICCV'25       & 1817          & 59.4          & \textbf{70.1}  & 85.8          & 57.2             & \textbf{51.7}   & 75.2            & 63.3            & 32.7            & 99.4\%              \\
SparseVLM         & ICML'25       & 1721          & 57.6          & 69.1           & 83.6          & 56.1             & 50.5            & 75.6            & 62.5            & 31.5            & 97.0\%              \\
HiRED             & AAAI'25       & 1737          & 58.7          & 68.4           & 82.8          & 47.4             & 50.1            & 74.9            & 62.8            & -               & 94.7\%              \\
DART              & EMNLP'25      & \textbf{1856} & 60.0          & 69.8           & 82.8          & 57.4             & 51.1            & 76.7            & 63.6            & 31.5            & 99.0\%              \\
VisionZIP         & CVPR'25       & 1783          & 59.3          & 68.9           & 85.3          & 57.3             & 50.9            & 76.8            & 63.0            & 31.7            & 98.5\%              \\
DivPrune          & CVPR'25        & 1751          & 60.0          & 68.7           & 87.0          & 56.4             & 51.2            & 75.5            & 62.3            & 32.0            & 98.4\%              \\
PDrop             & CVPR'25        & 1766          & 57.1          & 68.8           & 82.3          & 56.1             & 51.1            & 75.1            & 63.2            & 30.5            & 96.8\%              \\
SCOPE             & NeurIPS'25       & 1804          & 60.1          & 68.8           & 86.4          & 57.7             & -               & -               & 63.6            & 32.5            & 99.3\%              \\
PruneSID          & ICLR'26       & 1791          & 60.1          & 68.5           & 86.9          & 56.7             & -               & 76.8            & 63.7            & -               & 98.1\%              \\
LearnPruner       & ICLR'26       & 1820          & 60.3          & 68.5           & 86.7          & 57.3             & -               & 77.3            & 63.8            & -               & 98.6\%              \\
\rowcolor[HTML]{F0F4FF} 
\textbf{ACPruner} & \textbf{Ours}  & 1800          & \textbf{60.5} & 68.9           & \textbf{87.5} & \textbf{57.9}    & 51.5            & \textbf{77.6}   & \textbf{64.5}   & \textbf{33.0}   & \textbf{100.2\%}    \\
\midrule
\rowcolor[HTML]{F5F6F7}
\multicolumn{12}{c}{\cellcolor[HTML]{F5F6F7}Retain 128 visual tokens (pruning ratio = 77.8\%)}                                                                                                     \\
\midrule
FastV             & ECCV'24       & 1490          & 49.6          & 60.2           & 59.6          & 50.6             & 51.3            & 61.8            & 56.1            & 28.1            & 84.6\%              \\
PruMerge          & ICCV'25       & 1554          & 53.3          & 67.1           & 67.2          & 54.3             & 50.3            & 68.8            & 58.1            & -               & 89.5\%              \\
FasterVLM         & arXiv'24      & 1762          & 57.8          & 70.0           & 82.8          & 56.3             & 51.1            & 73.9            & 62.5            & 32.5            & 97.6\%              \\
TRIM              & COLING'25     & 1743          & 58.4          & 68.6           & 85.3          & 52.2             & 51.6            & 75.4            & 63.0            & 29.9            & 96.4\%              \\
VisPruner         & ICCV'25       & 1771          & 58.0          & 69.1           & 84.6          & 57.0             & \textbf{52.7}   & 73.9            & 61.9            & \textbf{33.7}   & 98.6\%              \\
SparseVLM         & ICML'25       & 1696          & 56.0          & 67.1           & 80.5          & 54.9             & 51.4            & 73.8            & 60.0            & 29.0            & 94.3\%              \\
HiRED             & AAAI'25       & 1710          & 57.2          & 68.1           & 79.8          & 46.1             & 51.3            & 73.4            & 61.5            & -               & 93.2\%              \\
DART              & EMNLP'25      & \textbf{1840} & 58.7          & 69.1           & 80.1          & 56.4             & 51.7            & 75.9            & 63.2            & 30.9            & 97.8\%              \\
VisionZIP         & CVPR'25       & 1762          & 57.6          & 68.9           & 83.2          & 56.8             & 51.6            & 75.6            & 62.0            & 32.6            & 97.9\%              \\
DivPrune          & CVPR'25       & 1752          & 59.2          & 69.0           & 86.9          & 56.0             & 52.7            & 74.7            & 62.3            & 30.7            & 97.9\%              \\
PDrop             & CVPR'25       & 1644          & 56.0          & 68.3           & 82.3          & 55.1             & 51.0            & 72.9            & 61.1            & 30.8            & 95.0\%              \\
SCOPE             & NeurIPS'25       & 1776          & 59.7          & 68.4           & 86.1          & 57.2             & -               & -               & 62.5            & 31.4            & 98.0\%              \\
AdaptPrune        & NeurIPS'25       & 1755          & 59.3          & 68.5           & 86.5          & 57.0             & 52.6            & 76.4            & 62.3            & -               & 98.3\%              \\
PruneSID          & ICLR'26       & 1749          & 58.8          & 68.3           & 86.5          & 54.7             & -               & 75.3            & 62.1            & -               & 96.3\%              \\
LearnPruner       & ICLR'26       & 1750          & 58.9          & 68.3           & 86.8          & 56.6             & -               & 76.0            & 62.6            & -               & 97.1\%              \\
\rowcolor[HTML]{F0F4FF} 
\textbf{ACPruner} & \textbf{Ours}  & 1812          & \textbf{59.8} & \textbf{69.4}  & \textbf{87.2} & \textbf{57.6}    & 52.0            & \textbf{76.8}   & \textbf{63.2}   & 32.8            & \textbf{99.8\%}     \\
\midrule
\rowcolor[HTML]{F5F6F7}
\multicolumn{12}{c}{\cellcolor[HTML]{F5F6F7}Retain 64 visual tokens (pruning ratio = 88.9\%)}                                                                                                      \\
\midrule
FastV             & ECCV'24       & 1256          & 46.1          & 51.1           & 48.0          & 47.8             & 50.8            & 55.0            & 48.0            & 26.7            & 76.1\%              \\
PruMerge          & ICCV'25       & 1549          & 51.9          & 68.1           & 65.3          & 54.0             & 50.1            & 67.4            & 55.3            & -               & 88.2\%              \\
FasterVLM         & arXiv'24      & 1667          & 55.0          & 70.2           & 76.6          & 55.3             & 51.7            & 70.6            & 60.6            & 31.5            & 94.6\%              \\
TRIM              & COLING'25     & 1680          & 56.6          & 69.0           & 85.9          & 49.7             & 51.1            & 72.4            & 60.9            & 24.8            & 92.7\%              \\
VisPruner         & ICCV'25       & 1691          & 55.4          & 69.1           & 80.4          & 55.8             & 53.3            & 70.9            & 60.1            & 31.7            & 95.6\%              \\
SparseVLM         & ICML'25       & 1505          & 52.7          & 62.2           & 75.1          & 51.8             & 50.1            & 68.2            & 56.2            & 24.9            & 87.3\%              \\
HiRED             & AAAI'25       & 1599          & 54.6          & 68.2           & 73.6          & 44.2             & 50.2            & 69.7            & 60.2            & -               & 89.5\%              \\
DART              & EMNLP'25      & \textbf{1765} & 55.9          & \textbf{69.8}  & 73.9          & 54.4             & 51.6            & 72.4            & 60.6            & 26.5            & 93.2\%              \\
VisionZIP         & CVPR'25       & 1690          & 55.1          & 69.0           & 77.0          & 55.5             & 52.9            & 72.4            & 60.1            & 31.7            & 95.2\%              \\
DivPrune          & CVPR'25       & 1638          & 57.6          & 68.3           & 85.6          & 55.5             & 53.6            & 72.9            & 59.3            & 29.4            & 95.6\%              \\
PDrop             & CVPR'25       & 1092          & 41.9          & 68.6           & 55.9          & 45.9             & 50.7            & 69.2            & 33.3            & 30.7            & 78.7\%              \\
SCOPE             & NeurIPS'25       & 1698          & 58.3          & 68.6           & 83.9          & 56.6             & -               & -               & 61.7            & 30.4            & 95.9\%              \\
AdaptPrune        & NeurIPS'25       & 1715          & 57.4          & 68.8           & 84.8          & 54.3             & \textbf{53.9}   & 74.7            & 61.4            & -               & 96.7\%              \\
PruneSID          & ICLR'26       & 1733          & 57.1          & 67.8           & 83.8          & 54.2             & -               & 73.7            & 58.8            & -               & 94.1\%              \\
LearnPruner       & ICLR'26       & 1672          & 57.2          & 68.2           & 84.5          & 56.1             & -               & 74.0            & 60.8            & -               & 94.8\%              \\
\rowcolor[HTML]{F0F4FF} 
\textbf{ACPruner} & \textbf{Ours}  & 1728          & \textbf{58.3} & 69.0           & \textbf{86.3} & \textbf{56.7}    & 53.1            & \textbf{75.1}   & \textbf{61.8}   & \textbf{32.0}   & \textbf{98.2\%}  
\\
\bottomrule
\end{tabular}}
\end{table}
\vspace{-4pt}

\subsection{Experimental Setup}
\label{experimental-setup}
\paragraph{Implementation details.}
We validate ACPruner on LVLMs from different model families and scales, including LLaVA-1.5-7B/13B~\cite{llava1.5}, LLaVA-NeXT-7B/13B~\cite{llavanext}, LLaVA-OneVision-7B~\cite{llavaOV}, and Qwen2.5-VL-7B~\cite{qwen2.5}. For LLaVA family, we set $\alpha=0.6$ and $\beta=1.0$. For Qwen2.5-VL, we set $\alpha=0.2$ and $\beta=1.0$. For noun extraction, we use the lightweight \texttt{en\_core\_web\_sm}~\cite{en_core_web_sm} model. 

\paragraph{Evaluation protocols.}
We compare ACPruner with pruning methods published at recent AI conferences across both image and video understanding tasks. For image understanding, we benchmark on GQA~\cite{gqa}, VizWiz~\cite{vizwiz}, VQA-v2~\cite{vqav2}, SQA~\cite{sqa}, TextVQA~\cite{textvqa}, POPE~\cite{pope}, MME~\cite{mme}, MMB~\cite{mmb}, and MM-Vet~\cite{mmvet}. For video understanding, we evaluate on MVBench~\cite{mvbench}, LongVideoBench~\cite{longvideobench}, NextQA~\cite{nextqa}, and VideoMME~\cite{videomme}. Details of the comparison methods and evaluation benchmarks are provided in Appendices~\ref{appendix:comparison} and~\ref{appendix:benchmarks}. Evaluations are conducted on a cluster of 8 NVIDIA L20 GPUs.

\subsection{Main Results}

\paragraph{Results on LLaVA-1.5-7B.}
LLaVA-1.5 adopts a fixed-resolution visual encoding strategy, where each input image is resized to $336^2$ and encoded into 576 visual tokens. Table~\ref{tab:llava-1.5-7b} demonstrates the performance comparison on LLaVA-1.5-7B under three pruning ratios. At the moderate pruning ratio of 66.7\%, ACPruner achieves the best overall performance retention of 100.2\%, even slightly surpassing the full-token baseline. This suggests that removing redundant visual tokens can help reduce input noise. At the more aggressive pruning ratio of 77.8\%, ACPruner remains the best-performing method overall, achieving 99.8\% performance retention. Even under the extremely high pruning ratio of 88.9\%, ACPruner still attains the highest average performance retention of 98.2\%.

\paragraph{Results on LLaVA-NeXT-7B.}
LLaVA-NeXT adopts an image tiling strategy for high-resolution visual encoding. Each input image is first resized to the closest aspect ratio in $\{$(336, 672), (672, 336), (672, 672), (1008, 336), (336, 1008)$\}$ and then partitioned into non-overlapping tiles, each of size $336^2$. Meanwhile, the original image is resized to $336^2$ as a global thumbnail. The local tiles and the global thumbnail are independently encoded by the vision encoder, yielding up to 2880 visual tokens in total. Table~\ref{tab:llava-next-7b} demonstrates the performance comparison on LLaVA-NeXT-7B under two pruning ratios. At the pruning ratio of 77.8\%, ACPruner achieves the best overall performance retention of 99.4\%, outperforming all comparison methods. At the more aggressive pruning ratio of 88.9\%, ACPruner still delivers the best average performance retention of 97.8\%. These results further demonstrate the robustness of ACPruner on high-resolution LVLMs.

\begin{table}[t]
\caption{Performance comparison on LLaVA-NeXT-7B under different pruning ratios. ``Perf. Ret.'' denotes the average performance retention rate relative to the full-token baseline.
}
\label{tab:llava-next-7b}
\resizebox{\columnwidth}{!}{
\begin{tabular}{l|c|cccccccc|c}
\toprule
\rowcolor[HTML]{F2F3F5} 
\textbf{Method}   & \textbf{Venue} & \textbf{MME}  & \textbf{GQA}  & \textbf{SQA} & \textbf{POPE} & \textbf{TextVQA} & \textbf{VizWiz} & \textbf{VQA-v2} & \textbf{MMB} & \textbf{Perf. Ret.} \\
\midrule
\rowcolor[HTML]{F5F6F7} 
\multicolumn{11}{c}{\cellcolor[HTML]{F5F6F7}Upper Bound, up to 2880 visual tokens  (pruning ratio = 0\%)}                                                                        \\
\midrule
LLaVA-NeXT-7B     & CVPR'24        & 1842          & 64.3          & 70.2           & 86.5          & 61.3             & 55.2            & 81.3            & 67.9            & 100.0\%             \\
\midrule
\rowcolor[HTML]{F5F6F7} 
\multicolumn{11}{c}{\cellcolor[HTML]{F5F6F7}Retain up to 640 visual tokens (pruning ratio  = 77.8\%)}                                                                            \\
\midrule
FastV             & ECCV'24        & 1807          & 58.9          & 67.4           & 79.5          & 58.1             & 53.9            & 77.0            & 63.1            & 94.7\%              \\
PruMerge          & ICCV'25        & 1790          & 60.8          & 67.8           & 85.3          & 54.9             & \textbf{57.9}   & 78.2            & 64.6            & 96.6\%              \\
DART              & EMNLP'25       & 1793          & 61.3          & 68.2           & 85.0          & 59.5             & 57.0            & 78.3            & 64.9            & 97.5\%              \\
SparseVLM         & ICML'25        & 1772          & 61.2          & 67.6           & 85.3          & 59.7             & 53.6            & 79.2            & 65.9            & 96.9\%              \\
PDrop             & CVPR'25        & 1782          & 60.0          & 66.7           & 83.8          & 57.8             & 53.8            & 79.1            & 64.1            & 95.7\%              \\
DivPrune          & CVPR'25        & 1773          & 61.9          & 67.8           & 86.9          & 57.0             & 55.7            & 79.3            & 65.8            & 97.2\%              \\
VisionZIP         & CVPR'25        & 1782          & 61.3          & 68.1           & 86.2          & 59.9             & 57.1            & 79.1            & 66.3            & 98.1\%              \\
SCOPE             & NeurIPS'25     & \textbf{1842} & 61.9          & 67.8           & -             & 60.1             & -               & -               & 66.2            & 97.7\%              \\
PruneSID          & ICLR'26        & 1795          & 61.6          & 68.3           & 86.3          & -                & -               & -               & 64.2            & 97.0\%              \\
AgilePruner       & ICLR'26        & 1802          & 62.0          & 67.8           & 86.1          & 59.0             & 56.0            & 79.3            & 65.9            & 97.8\%              \\
\rowcolor[HTML]{F0F4FF} 
\textbf{ACPruner} & \textbf{Ours}  & 1821          & \textbf{63.2} & \textbf{69.0}  & \textbf{87.4} & \textbf{60.2}    & 57.2            & \textbf{79.6}   & \textbf{67.1}   & \textbf{99.4\%}     \\
\midrule
\rowcolor[HTML]{F5F6F7} 
\multicolumn{11}{c}{\cellcolor[HTML]{F5F6F7}Retain up to 320 visual tokens (pruning ratio  = 88.9\%)}                                                                            \\
\midrule
FastV             & ECCV'24        & 1539          & 49.8          & 66.6           & 49.5          & 52.2             & 51.3            & 61.5            & 53.4            & 80.7\%              \\
PruMerge          & ICCV'25        & 1744          & 58.8          & 68.1           & 79.5          & 54.0             & \textbf{57.7}   & 75.3            & 63.0            & 94.1\%              \\
DART              & EMNLP'25       & 1710          & 59.5          & 67.5           & 81.0          & 57.6             & 56.1            & 75.7            & 64.2            & 94.8\%              \\
SparseVLM         & ICML'25        & 1747          & 57.9          & 67.2           & 76.9          & 56.5             & 54.2            & 74.6            & 63.1            & 93.1\%              \\
PDrop             & CVPR'25        & 1672          & 50.4          & 66.7           & 60.8          & 49.0             & 49.7            & 66.8            & 55.5            & 83.5\%              \\
DivPrune          & CVPR'25        & 1731          & 61.1          & 67.7           & 84.7          & 56.2             & 55.6            & 77.2            & 63.9            & 95.6\%              \\
VisionZIP         & CVPR'25        & 1698          & 59.3          & 67.3           & 82.1          & 58.9             & 56.2            & 76.2            & 63.1            & 95.0\%              \\
SCOPE             & NeurIPS'25     & \textbf{1789} & 61.0          & 67.7           & -             & 58.4             & -               & -               & \textbf{65.9}   & 96.2\%              \\
PruneSID          & ICLR'26        & 1754          & 60.5          & 67.3           & 83.1          & -                & -               & -               & 63.0            & 94.8\%              \\
AgilePruner       & ICLR'26        & 1760          & 60.1          & 67.3           & 84.0          & 58.4             & 55.8            & 77.8            & 64.5            & 96.2\%              \\
\rowcolor[HTML]{F0F4FF} 
\textbf{ACPruner} & \textbf{Ours}  & 1774          & \textbf{62.4} & \textbf{68.8}  & \textbf{87.1} & \textbf{59.1}    & 56.5            & \textbf{78.2}   & 65.0            & \textbf{97.8\%}    \\
\bottomrule
\end{tabular}}
\end{table}

\begin{table}[h]
\caption{Performance comparison on LLaVA-OneVision-7B for video understanding tasks. ``Perf. Ret.'' denotes the average performance retention rate relative to the full-token baseline.
}
\label{tab:llava-ov-7b}
\resizebox{\columnwidth}{!}{
\begin{tabular}{l|c|cccccccc|c}
\toprule
\rowcolor[HTML]{F2F3F5} 
\cellcolor[HTML]{F2F3F5}                                  & \cellcolor[HTML]{F2F3F5}                                 & \cellcolor[HTML]{F2F3F5}                                   & \cellcolor[HTML]{F2F3F5}                                     & \cellcolor[HTML]{F2F3F5}                                  & \multicolumn{4}{c}{\cellcolor[HTML]{F2F3F5}\textbf{VideoMME}}       & \cellcolor[HTML]{F2F3F5}                                & \cellcolor[HTML]{F2F3F5}                                      \\
\multirow{-2}{*}{\cellcolor[HTML]{F2F3F5}\textbf{Method}} & \multirow{-2}{*}{\cellcolor[HTML]{F2F3F5}\textbf{Venue}} & \multirow{-2}{*}{\cellcolor[HTML]{F2F3F5}\textbf{MVBench}} & \multirow{-2}{*}{\cellcolor[HTML]{F2F3F5}\textbf{LongVideo}} & \multirow{-2}{*}{\cellcolor[HTML]{F2F3F5}\textbf{NextQA}} & \cellcolor[HTML]{F2F3F5}Overall & \cellcolor[HTML]{F2F3F5}Short & \cellcolor[HTML]{F2F3F5}Medium & \cellcolor[HTML]{F2F3F5}Long & \multirow{-2}{*}{\cellcolor[HTML]{F2F3F5}\textbf{AVG.}} & \multirow{-2}{*}{\cellcolor[HTML]{F2F3F5}\textbf{Perf. Ret.}} \\
\midrule

\rowcolor[HTML]{F5F6F7} 
\multicolumn{11}{c}{\cellcolor[HTML]{F5F6F7} Pruning Ratio = 0\%}                                                                                                                                                                                                                                                                                                                                                                                                                \\
\midrule

\cellcolor[HTML]{F2F3F5}\textbf{LLaVA-OV-7B}              & TMLR'25                                                  & 56.9                                                       & 56.4                                                         & 79.0                                                      & 58.6             & 70.3           & 56.6            & 48.8          & 62.7                                                    & 100.0\%                                                       \\
\midrule
\rowcolor[HTML]{F5F6F7} 
\multicolumn{11}{c}{\cellcolor[HTML]{F5F6F7} Pruning Ratio = 75\%}                                                                                                                                                                                                                                                                                                                                                                                                                \\
\midrule
\cellcolor[HTML]{F2F3F5}\textbf{FastV}                    & ECCV'24                                                  & 54.7                                                       & 55.5                                                         & 77.5                                                      & 56.2             & 68.0           & 54.6            & 46.0          & 61.0                                                    & 97.1\%                                                        \\
\cellcolor[HTML]{F2F3F5}\textbf{DivPrune}                 & CVPR'25                                                  & 54.6                                                       & 55.7                                                         & 77.7                                                      & 57.1             & \textbf{69.0}  & 54.6            & 47.9          & 61.3                                                    & 97.6\%                                                        \\
\cellcolor[HTML]{F2F3F5}\textbf{VisionZIP}                & CVPR'25                                                  & 53.7                                                       & 51.2                                                         & 77.6                                                      & 54.1             & 61.6           & 53.4            & 47.2          & 59.2                                                    & 93.9\%                                                        \\
\cellcolor[HTML]{F2F3F5}\textbf{PDrop}                    & CVPR'25                                                  & 55.3                                                       & 54.1                                                         & 74.2                                                      & 55.5             & 64.7           & 53.1            & \textbf{48.7} & 59.8                                                    & 95.5\%                                                        \\
\cellcolor[HTML]{F2F3F5}\textbf{SparseVLM}                & ICML'25                                                  & 56.4                                                       & 53.9                                                         & 76.2                                                      & 57.3             & 68.4           & 55.2            & 48.1          & 61.0                                                    & 97.2\%                                                        \\
\rowcolor[HTML]{F0F4FF} 
\textbf{ACPruner}                                         & \textbf{Ours}                                            & \textbf{57.0}                                              & \textbf{55.9}                                                & \textbf{77.7}                                             & \textbf{57.6}    & 68.6           & \textbf{55.8}   & 48.3          & \textbf{62.1}                                           & \textbf{99.0\%}                                               \\
\midrule
\rowcolor[HTML]{F5F6F7} 
\multicolumn{11}{c}{\cellcolor[HTML]{F5F6F7} Pruning Ratio = 85\%}                                                                                                                                                                                                                                                                                                                                                                                                                                 \\
\midrule
\cellcolor[HTML]{F2F3F5}\textbf{FastV}                    & ECCV'24                                                  & 53.2                                                       & 54.9                                                         & 77.1                                                      & 54.7             & 65.1           & 53.4            & 45.0          & 60.0                                                    & 95.4\%                                                        \\
\cellcolor[HTML]{F2F3F5}\textbf{DivPrune}                 & CVPR'25                                                  & \textbf{55.9}                                              & 55.1                                                         & 76.8                                                      & 56.3             & 66.7           & 54.2            & \textbf{48.2} & 61.0                                                    & 97.4\%                                                        \\
\cellcolor[HTML]{F2F3F5}\textbf{VisionZIP}                & CVPR'25                                                  & 44.4                                                       & 43.5                                                         & 77.2                                                      & 46.0             & 50.4           & 45.8            & 41.8          & 52.8                                                    & 82.8\%                                                        \\
\cellcolor[HTML]{F2F3F5}\textbf{PDrop}                    & CVPR'25                                                  & 53.2                                                       & 47.6                                                         & 74.5                                                      & 50.1             & 58.7           & 48.7            & 45.0          & 56.4                                                    & 89.4\%                                                        \\
\cellcolor[HTML]{F2F3F5}\textbf{SparseVLM}                & ICML'25                                                  & 52.9                                                       & 54.2                                                         & 76.6                                                      & 52.9             & 62.6           & 51.1            & 44.6          & 59.2                                                    & 94.1\%                                                        \\
\rowcolor[HTML]{F0F4FF} 
\textbf{ACPruner}                                         & \textbf{Ours}                                            & 55.8                                                       & \textbf{56.2}                                                & \textbf{78.2}                                             & \textbf{56.5}    & \textbf{66.7}  & \textbf{54.9}   & 47.9          & \textbf{61.7}                                           & \textbf{98.3\%}                                                       
\\
\bottomrule
\end{tabular}
}
\end{table}

\paragraph{Results on Qwen2.5-VL-7B.}

\setlength{\tabcolsep}{2pt}
\begin{wraptable}{r}{0.6\columnwidth}
\vspace{-0.8em}
\caption{Performance comparison on Qwen2.5-VL-7B under different pruning ratios. ``Perf. Ret.'' denotes the average performance retention rate relative to the full-token baseline.}
\label{tab:qwen-2.5-vl-7b}
\raggedright
\resizebox{\linewidth}{!}{%
\begin{tabular}{l|ccccccc|c}
\toprule
\rowcolor[HTML]{F2F3F5} 
\textbf{Method } & \textbf{MME}  & \textbf{POPE} & \textbf{TextVQA} & \textbf{MMB} & \textbf{GQA}  & \textbf{SQA}  & \textbf{VQA-v2} & \textbf{Perf. Ret.} \\
\midrule
\rowcolor[HTML]{F5F6F7} 
\multicolumn{9}{c}{\cellcolor[HTML]{F5F6F7} Pruning ratio = 0\%} \\
\midrule
Qwen2.5-VL-7B & 2310 & 86.3 & 84.8 & 83.9 & 60.9 & 88.9 & 82.9 & 100.0\% \\
\midrule
\rowcolor[HTML]{F5F6F7} 
\multicolumn{9}{c}{\cellcolor[HTML]{F5F6F7} Pruning ratio = 66.7\%} \\
\midrule
FastV      & 2072 & 82.2 & 77.9 & 75.7 & 58.0 & 78.5 & 80.4 & 92.5\% \\
DivPrune   & 2198 & 85.6 & 80.1 & 81.6 & 59.0 & 86.6 & 80.9 & 96.8\% \\
VisionZIP  & 2317 & 85.8 & 80.4 & 78.9 & 56.6 & 80.5 & 80.7 & 95.6\% \\
PruneSID   & 2218 & 85.9 & 80.8 & 80.9 & 59.8 & \textbf{87.6} & 80.4 & 97.3\% \\
\rowcolor[HTML]{F0F4FF} 
\textbf{ACPruner} & \textbf{2319} & \textbf{87.4} & \textbf{81.0} & \textbf{81.9} & \textbf{60.0} & 87.3 & \textbf{81.2} & \textbf{98.5\%} \\
\midrule
\rowcolor[HTML]{F5F6F7} 
\multicolumn{9}{c}{\cellcolor[HTML]{F5F6F7} Pruning ratio = 77.8\%} \\
\midrule
FastV      & 2036 & 80.7 & 69.0 & 74.9 & 56.7 & 78.0 & 78.1 & 89.6\% \\
DivPrune   & 2153 & 85.5 & 76.6 & 78.6 & 58.6 & 86.2 & 78.8 & 94.9\% \\
VisionZIP  & 2224 & 83.4 & 74.5 & 76.8 & 54.6 & 80.4 & 78.5 & 92.4\% \\
PruneSID   & 2169 & 85.6 & 77.0 & 78.0 & 59.0 & 86.9 & 78.7 & 95.2\% \\
\rowcolor[HTML]{F0F4FF} 
\textbf{ACPruner} & \textbf{2302} & \textbf{86.9} & \textbf{77.9} & \textbf{79.4} & \textbf{59.2} & \textbf{87.3} & \textbf{79.2} & \textbf{96.8\%} \\
\midrule
\rowcolor[HTML]{F5F6F7} 
\multicolumn{9}{c}{\cellcolor[HTML]{F5F6F7} Pruning ratio = 88.9\%} \\
\midrule
FastV      & 1940 & 78.6 & 60.3 & 69.2 & 51.9 & 77.4 & 55.2 & 81.1\% \\
DivPrune   & 2051 & 83.7 & 66.2 & 76.4 & 56.9 & 82.1 & 74.5 & 90.1\% \\
VisionZIP  & 2025 & 78.9 & 63.3 & 75.8 & 53.2 & 80.1 & 73.8 & 87.2\% \\
PruneSID   & 2076 & 80.2 & 66.5 & 73.9 & 55.8 & \textbf{86.5} & 74.6 & 89.7\% \\
\rowcolor[HTML]{F0F4FF} 
\textbf{ACPruner} & \textbf{2189} & \textbf{85.5} & \textbf{69.0} & \textbf{76.9} & \textbf{56.8} & 82.6 & \textbf{75.1} & \textbf{91.4\%} \\
\bottomrule
\end{tabular}}
\vspace{-1em}
\end{wraptable}

As a more advanced LVLM family, Qwen2.5-VL directly trains its vision encoder from scratch within the LVLM framework. This design enables the encoder to natively process images at varying resolutions, producing a dynamic number of visual tokens that adapt to the input image size. Table~\ref{tab:qwen-2.5-vl-7b} demonstrates the performance comparison on Qwen2.5-VL-7B under three pruning ratios. At the pruning ratio of 66.7\%, ACPruner achieves the best overall performance retention of 98.5\%, outperforming all comparison methods. At the more aggressive pruning ratio of 77.8\%, ACPruner still delivers the highest average performance retention of 96.8\%. Even under the extremely high pruning ratio of 88.9\%, ACPruner continues to achieve the best overall performance retention of 91.4\%. 
These results show that ACPruner generalizes well to more advanced LVLMs and remains effective under larger and highly variable visual token counts.

\paragraph{Results on Video Understanding.}
Following prior work, we adopt LLaVA-OneVision-7B as the backbone for video understanding, where each video is uniformly sampled into 32 frames and each frame is encoded into 196 visual tokens. Table~\ref{tab:llava-ov-7b} shows the performance comparison on four widely used video understanding benchmarks under two pruning ratios. At the pruning ratio of 75\%, ACPruner achieves the best overall performance retention of 99.0\%, outperforming all comparison methods. At the more aggressive pruning ratio of 85\%, ACPruner still delivers the best average performance retention of 98.3\%. These results show that ACPruner remains effective not only for image understanding, but also for temporally extended video understanding tasks.

\paragraph{Results on Larger-Scale LVLMs.}
We further evaluate ACPruner on larger-scale LVLMs, including LLaVA-1.5-13B and LLaVA-NeXT-13B. As shown in Appendix~\ref{results-13b}, ACPruner consistently remains the best-performing method across different pruning ratios, showing that the proposed formulation generalizes well not only across LVLM families, but also across model scales.

\subsection{Practical Efficiency Analysis}
\label{efficiency}
We report both benchmark-level efficiency statistics and sample-level runtime breakdowns in Table~\ref{tab:efficiency}. All efficiency experiments are conducted on a single NVIDIA L20 GPU. At the benchmark level, ACPruner consistently yields substantial efficiency gains across the evaluated LVLM backbones. Two trends are particularly evident. First, under the same model scale, high-resolution LVLMs benefit more from visual token pruning because they start with much longer visual token sequences. Second, under the same family, larger models achieve greater practical speedups because shortening the visual sequence leads to larger savings in the more expensive LLM inference stage. The sample-level runtime breakdown further shows that the pruning overhead remains modest and does not become an efficiency bottleneck. Across all backbones, the additional cost introduced by ACPruner is much smaller than the runtime reduction in downstream LLM inference, indicating that the speedup mainly comes from reducing the computation within the LLM, rather than being offset by the pruning procedure itself. The runtime breakdown of ACPruner's internal steps is provided in Appendix~\ref{appendix:runtime}.

\begin{table}[]
\caption{Practical efficiency analysis of ACPruner on MME. The left section shows benchmark-level efficiency statistics while the right section shows sample-level runtime breakdowns.}
\label{tab:efficiency}
\resizebox{\columnwidth}{!}{
\begin{tabular}{l|ccccc|cccc}
\toprule
                                  &                                                             &                                                                                   &                                                                                    &                                                                                           &                                        & \multicolumn{4}{c}{\textbf{Sample-level Inference Time Breakdown (MilliSeconds)}} \\
\multirow{-2}{*}{\textbf{Method}} & \multirow{-2}{*}{\textbf{\# Tokens}}                        & \multirow{-2}{*}{\textbf{\begin{tabular}[c]{@{}c@{}}FLOPs\\ (Tera)\end{tabular}}} & \multirow{-2}{*}{\textbf{\begin{tabular}[c]{@{}c@{}}KV-cache\\ (MB)\end{tabular}}} & \multirow{-2}{*}{\textbf{\begin{tabular}[c]{@{}c@{}}Eval. Time\\ (Seconds)\end{tabular}}} & \multirow{-2}{*}{\textbf{Score}} & Visual Encoding        & Token Pruning        & LLM Inference        & Total       \\
\midrule
LLaVA-1.5-7B                      & 576                                                         & 8.5                                                                               & 318.6                                                                              & 511.8                                                                                     & 1862                                   & 14.6                   & -                    & 201.4               & 215.7       \\
\rowcolor[HTML]{F0F4FF} 
\textbf{+ Ours}                   & 64 (\textcolor{green!50!black}{$\downarrow 88.9\%$})  & 1.6 (\textcolor{green!50!black}{$\downarrow 81.2\%$})                       & 62.6 (\textcolor{green!50!black}{$\downarrow 80.4\%$})                       & 299.1 (\textcolor{green!50!black}{$\downarrow 41.6\%$})                             & 1728                                   & 14.6                   & 10.2                 & 101.3               & 126.1       \\
LLaVA-1.5-13B                     & 576                                                         & 16.5                                                                              & 497.9                                                                              & 742.9                                                                                     & 1818                                   & 14.6                   & -                    & 298.5               & 313.1       \\
\rowcolor[HTML]{F0F4FF} 
\textbf{+ Ours}                   & 64 (\textcolor{green!50!black}{$\downarrow 88.9\%$})  & 3.2 (\textcolor{green!50!black}{$\downarrow 80.6\%$})                       & 97.9 (\textcolor{green!50!black}{$\downarrow 80.4\%$})                       & 380.3 (\textcolor{green!50!black}{$\downarrow 48.8\%$})                             & 1820                                   & 14.6                   & 10.5                 & 135.2               & 160.3       \\
LLaVA-NeXT-7B                     & 2880                                                        & 30.6                                                                              & 1084.7                                                                             & 1184.4                                                                                    & 1842                                   & 35.9                   & -                    & 463.2               & 499.1       \\
\rowcolor[HTML]{F0F4FF} 
\textbf{+ Ours}                    & 320 (\textcolor{green!50!black}{$\downarrow 88.9\%$}) & 4.5 (\textcolor{green!50!black}{$\downarrow 85.3\%$})                       & 170.6 (\textcolor{green!50!black}{$\downarrow 84.3\%$})                      & 683.3 (\textcolor{green!50!black}{$\downarrow 42.3\%$})                             & 1774                                   & 35.9                   & 19.8                 & 232.3               & 287.9       \\
LLaVA-NeXT-13B                    & 2880                                                        & 58.9                                                                              & 1694.9                                                                             & 1853.8                                                                                    & 1901                                   & 35.9                   & -                    & 745.3               & 781.2       \\
\rowcolor[HTML]{F0F4FF} 
\textbf{+ Ours}                    & 320 (\textcolor{green!50!black}{$\downarrow 88.9\%$}) & 8.8 (\textcolor{green!50!black}{$\downarrow 85.1\%$})                       & 266.6 (\textcolor{green!50!black}{$\downarrow 84.3\%$})                      & 868.2 (\textcolor{green!50!black}{$\downarrow 53.2\%$})                             & 1848                                   & 35.9                   & 20.3                 & 309.7               & 365.9      \\
\bottomrule
\end{tabular}}
\end{table}

\subsection{Ablation Study}
\begin{figure}[t]
    \centering
    \includegraphics[width=\columnwidth]{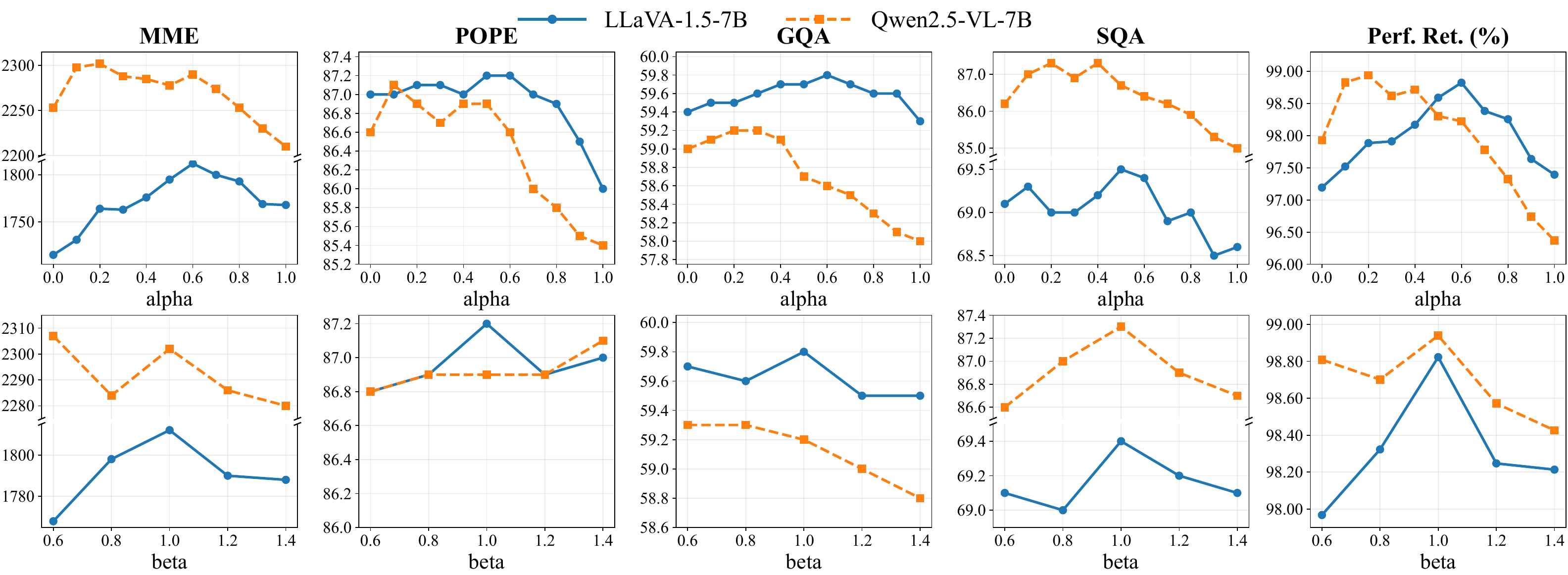}
    \caption{Ablation of $\alpha$ and $\beta$ on LLaVA-1.5-7B and Qwen2.5-VL-7B under a 77.8\% pruning ratio.}
    \label{fig:ablation_alpha_beta}
\end{figure}

\paragraph{Main Component Ablation.}

\setlength{\tabcolsep}{2pt}
\begin{wraptable}{r}{0.5\columnwidth}
\vspace{-0.5em}
\caption{Ablation of ACPruner's main components on LLaVA-1.5-7B under a 77.8\% pruning ratio.}
\label{tab:main_ablation}
\resizebox{0.5\columnwidth}{!}{
\begin{tabular}{l|ccccc|c}
\toprule
\rowcolor[HTML]{F2F3F5}
\textbf{}   & \textbf{MME} & \textbf{GQA} & \textbf{SQA} & \textbf{POPE} & \textbf{TextVQA} & \textbf{Perf. Ret.} \\
\midrule
\rowcolor[HTML]{F0F4FF}
Full method & 1812 & 59.8 & 69.4 & 87.2 & 57.6 & 98.9\% \\
Setting-1   & 1768 & 59.3 & 68.6 & 86.0 & 57.4 & 97.6\% \\
Setting-2   & 1715 & 59.4 & 69.1 & 87.0 & 54.6 & 96.5\% \\
Setting-3   & 1700 & 59.3 & 68.7 & 86.3 & 50.9 & 94.8\% \\
Setting-4   & 1780 & 57.9 & 68.3 & 84.8 & 53.3 & 95.5\% \\
Setting-5   & 1749 & 59.0 & 68.6 & 86.8 & 53.1 & 96.0\% \\
Random      & 1701 & 57.7 & 68.7 & 81.6 & 33.0 & 87.0\% \\
\bottomrule
\end{tabular}}
\vspace{-1em}
\end{wraptable}

Table~\ref{tab:main_ablation} presents the ablation results of the main components in ACPruner. In each setting, we remove a single component while keeping the remaining design unchanged. Removing inter-modality importance (setting-1) or intra-modality importance (setting-2) leads to clear performance drops, indicating that the two importance signals provide complementary cues. Removing the importance bias entirely (setting-3) causes further performance degradation, highlighting the necessity of prioritizing critical content during coverage maximization. When coverage modeling is discarded and token selection reduces to pure importance-based TopK selection (setting-4), the performance also drops noticeably, confirming the benefit of preserving the original token set collectively rather than selecting tokens solely by individual importance. Replacing attention-induced coverage with feature similarity (setting-5) further degrades performance, supporting our claim that patch-to-patch attention provides a more suitable relation for modeling token preservation. Additional ablation results on detailed design choices are provided in Appendix~\ref{appendix-ablation}.

\paragraph{Hyperparameter Ablation.}
Fig.~\ref{fig:ablation_alpha_beta} presents the ablation results of $\alpha$ and $\beta$ on both LLaVA-1.5-7B and Qwen2.5-VL-7B. For $\alpha$, LLaVA-1.5-7B achieves the best overall performance around $\alpha=0.6$, while Qwen2.5-VL-7B performs best around $\alpha=0.2$. This difference may be attributed to the different intra-modality importance signals used by the two backbones: CLS-to-patch attention is used for LLaVA-1.5 and patch incoming attention is used for Qwen2.5-VL. For $\beta$, both backbones obtain the strongest overall performance around $\beta=1.0$, indicating that ACPruner benefits from a balanced weighting between content importance and attention-based coverage. Overall, these results support the hyperparameter settings adopted in our experiments.

\section{Related Work: Visual Token Pruning in LVLMs}

While LVLMs have achieved strong multimodal understanding capabilities~\cite{llava1.5,llavanext,qwen2.5,llavaOV}, the large number of visual tokens introduces substantial inference overhead. Since recent studies have revealed that visual tokens in LVLMs are highly redundant~\cite{fastv,fastervlm,dart,prumerge}, developing training-free visual token pruning strategies has become an important direction for efficient LVLM inference.

One representative line of work uses crossmodal attention in the LLM to retain the most textually relevant visual tokens~\cite{fastv,sparsevlm,pdrop}. While these methods are text-aware and empirically effective, they require forwarding all visual tokens through early LLM layers before pruning, and their reliance on LLM attention signals usually makes them incompatible with FlashAttention~\cite{flashattention}. Moreover, recent studies have shown that crossmodal attention in early LLM layers exhibits significant positional bias, which can harm pruning performance~\cite{dart,arewesolving}. Another line uses CLS attention in the vision encoder to retain the most visually salient tokens before entering the LLM~\cite{fastervlm,hired,vispruner,prumerge,visionzip}. These methods are efficient because pruning is performed in a pre-LLM manner, but they are usually text-agnostic and may repeatedly preserve redundant tokens around salient regions. A third line focuses on reducing redundancy by selecting diverse tokens according to feature similarity in the embedding space~\cite{divprune,dart}. Although such methods can preserve broader image semantics, they may allocate token budgets to uninformative regions and are also typically agnostic to the input text.

Compared with prior work, ACPruner simultaneously achieves three desirable properties. First, it operates in a pre-LLM manner, effectively reducing the computation of LLM inference. Second, it is redundancy-aware: during greedy selection, tokens whose content has already been well preserved by the current subset provide smaller marginal gains and are therefore less likely to be selected. Third, it is importance-aware, since the coverage objective gives higher priority to preserving important visual content. Moreover, the importance estimation is text-aware, as it incorporates textual relevance, allowing ACPruner to produce query-adaptive pruning results (as shown in Appendix~\ref{appendix:case_study}).

\section{Conclusion and Limitations}
\label{conclusion}
In this work, we formulate visual token pruning as a biased attention coverage maximization problem and propose \textbf{ACPruner}, a training-free and pre-LLM visual token pruning framework for efficient LVLM inference. Unlike prior methods, ACPruner selects a compact token subset that collectively preserves the original visual information while assigning higher priority to more informative content. Experiments across diverse model families demonstrate that ACPruner achieves strong performance retention with substantial efficiency gains.
Despite its effectiveness, ACPruner has two potential limitations. First, it relies on two manually selected hyperparameters, $\alpha$ and $\beta$, whose optimal values may vary across tasks and backbones. While the default settings work well on public benchmarks, applying ACPruner to new scenarios may require hyperparameter re-tuning. Second, ACPruner assumes that the backbone model adopts a ViT-based vision encoder, whose attention signals are used to define both intra-modality token importance and token-to-token coverage. Although most LVLMs rely on ViT-based vision encoders, this architectural dependence may limit the applicability of ACPruner to LVLMs that do not employ an explicit vision encoder~\cite{eve, evev2}. Extending the biased coverage formulation to such architectures is an important direction for future work.

{
\small
\bibliographystyle{plainnat}
\bibliography{references}
}


\appendix

\section{Appendix Overview}
\label{appendix:overview}

This appendix provides additional theoretical analysis and experimental results of ACPruner. Appendix~\ref{appendix:theory} presents the theoretical analysis of the proposed biased attention coverage objective. Appendix~\ref{appendix:nphard} proves the NP-hardness of the proposed objective. Appendix~\ref{appendix:pseudocode} provides the complete pseudocode of ACPruner. Appendix~\ref{results-13b} reports additional experiments on larger-scale LVLMs, including LLaVA-1.5-13B and LLaVA-NeXT-13B. Appendix~\ref{appendix-ablation} provides additional ablation studies on design choices. Appendix~\ref{appendix:runtime} analyzes the step-wise runtime distribution. Appendix~\ref{appendix:benchmarks} introduces the image and video benchmarks used in our experiments. Appendix~\ref{appendix:comparison} summarizes the comparison methods included in the main experiments. Appendix~\ref{appendix:case_study} presents qualitative pruning visualizations.

\section{Theoretical Analysis of ACPruner}
\label{appendix:theory}

In this section, we show that the proposed objective admits a well-structured monotone submodular form, which justifies the use of greedy optimization and clarifies the role of jointly modeling importance and coverage.

\subsection{Preliminaries and Notation}

To simplify notation, we absorb the importance bias exponent into the content weight and define
\begin{equation}
a_j := w_j^{\beta}, \qquad a_j \ge 0,\quad j=1,\ldots,T.
\label{eq:appendix_weight}
\end{equation}
Then the ACPruner objective in Eq.~\eqref{eq:final_objective} can be written as
\begin{equation}
F(\mathbf{S})
:=
\sum_{j=1}^{T} a_j\, \mathrm{Cov}(\mathbf{S}, \mathbf{v}_j)
=
\sum_{j=1}^{T} a_j \max_{\mathbf{v}_i\in \mathbf{S}} c_{ij},
\qquad |\mathbf{S}| \le k,
\label{eq:appendix_objective}
\end{equation}
where, by definition,
\begin{equation}
\mathrm{Cov}(\mathbf{S}, \mathbf{v}_j)
=
\begin{cases}
\max_{\mathbf{v}_i \in \mathbf{S}} c_{ij}, & \text{if } \mathbf{S}\neq\emptyset,\\
0, & \text{if } \mathbf{S}=\emptyset.
\end{cases}
\label{eq:appendix_cov_def}
\end{equation}

Although the main text writes the constraint as $|\mathbf{S}|=k$, we use $|\mathbf{S}|\le k$ here for convenience. As will be shown below, $F(\mathbf{S})$ is monotone non-decreasing, and therefore the optimal values under $|\mathbf{S}|=k$ and $|\mathbf{S}|\le k$ are identical.

For notational convenience, we define the current coverage state of token $\mathbf{v}_j$ under subset $\mathbf{S}$ as
\begin{equation}
m_j(\mathbf{S}) := \mathrm{Cov}(\mathbf{S}, \mathbf{v}_j).
\label{eq:appendix_mj}
\end{equation}
Then Eq.~\eqref{eq:appendix_objective} becomes
\begin{equation}
F(\mathbf{S})=\sum_{j=1}^{T} a_j\, m_j(\mathbf{S}).
\label{eq:appendix_objective_compact}
\end{equation}

Since each attention weight satisfies $A_{i\rightarrow j}^{(l,h)}\in[0,1]$, their average also satisfies
\begin{equation}
0 \le c_{ij} \le 1, \qquad \forall i,j.
\label{eq:appendix_cij_range}
\end{equation}
Consequently,
\begin{equation}
0 \le m_j(\mathbf{S}) \le 1, \qquad \forall j,\ \forall \mathbf{S}\subseteq \mathbf{V}.
\label{eq:appendix_mj_range}
\end{equation}

For a candidate token $\mathbf{v}_u \in \mathbf{V}\setminus \mathbf{S}$, we denote its marginal gain with respect to $\mathbf{S}$ as
\begin{equation}
\Delta(\mathbf{v}_u \mid \mathbf{S})
:=
F(\mathbf{S}\cup\{\mathbf{v}_u\}) - F(\mathbf{S}).
\label{eq:appendix_marginal_def}
\end{equation}

\subsection{ACPruner as Weighted Distortion Minimization}

We first show that maximizing the ACPruner objective is equivalent to minimizing an importance-weighted representative distortion. This provides a principled interpretation of ACPruner as selecting a compact representative subset for the original visual token set.

\paragraph{Proposition 1.}
Define the distortion of token $\mathbf{v}_j$ under subset $\mathbf{S}$ as
\begin{equation}
d(\mathbf{v}_j,\mathbf{S}) := 1 - \mathrm{Cov}(\mathbf{S}, \mathbf{v}_j)
= 1 - m_j(\mathbf{S}).
\label{eq:appendix_distortion}
\end{equation}
Then maximizing the ACPruner objective in Eq.~\eqref{eq:appendix_objective} is equivalent to minimizing the weighted total distortion
\begin{equation}
D(\mathbf{S}) := \sum_{j=1}^{T} a_j\, d(\mathbf{v}_j,\mathbf{S}).
\label{eq:appendix_total_distortion}
\end{equation}
More precisely,
\begin{equation}
D(\mathbf{S}) = \sum_{j=1}^{T} a_j - F(\mathbf{S}).
\label{eq:appendix_distortion_equivalence}
\end{equation}

\paragraph{Proof.}
Substituting Eq.~\eqref{eq:appendix_distortion} into Eq.~\eqref{eq:appendix_total_distortion}, we obtain
\begin{align}
D(\mathbf{S})
&=
\sum_{j=1}^{T} a_j \left(1 - m_j(\mathbf{S})\right) \notag\\
&=
\sum_{j=1}^{T} a_j - \sum_{j=1}^{T} a_j\, m_j(\mathbf{S}) \notag\\
&=
\sum_{j=1}^{T} a_j - F(\mathbf{S}),
\end{align}
where the last equality follows from Eq.~\eqref{eq:appendix_objective_compact}. Since $\sum_{j=1}^{T} a_j$ is independent of $\mathbf{S}$, maximizing $F(\mathbf{S})$ is equivalent to minimizing $D(\mathbf{S})$. This proves the proposition.
\hfill $\square$

Proposition~1 shows that ACPruner is not merely selecting tokens with high individual scores. Instead, it seeks a compact subset that minimizes the importance-weighted representation error over the full token set, where the representative relation is induced by encoder attention.

\subsection{Normalization, Monotonicity, and Submodularity}

We next prove that the ACPruner objective is a normalized monotone submodular set function. This result explains why the greedy optimization used by ACPruner is mathematically well-founded.

\paragraph{Theorem 1.}
The set function
\begin{equation}
F(\mathbf{S})=\sum_{j=1}^{T} a_j\, m_j(\mathbf{S})
\label{eq:appendix_theorem1_objective}
\end{equation}
is normalized, monotone, and submodular.

\paragraph{Proof.}
We prove the three properties one by one.

\paragraph{(i) Normalization.}
By Eq.~\eqref{eq:appendix_cov_def}, for the empty set $\emptyset$ we have
\[
m_j(\emptyset)=\mathrm{Cov}(\emptyset,\mathbf{v}_j)=0,\qquad \forall j.
\]
Therefore,
\begin{equation}
F(\emptyset)=\sum_{j=1}^{T} a_j\, m_j(\emptyset)=0.
\label{eq:appendix_normalized}
\end{equation}
Hence $F$ is normalized.

\paragraph{(ii) Monotonicity.}
Let $\mathbf{S}\subseteq \mathbf{T}\subseteq \mathbf{V}$. Then for every $j$,
\begin{equation}
m_j(\mathbf{S})=\max_{\mathbf{v}_i\in\mathbf{S}} c_{ij}
\le
\max_{\mathbf{v}_i\in\mathbf{T}} c_{ij}
= m_j(\mathbf{T}),
\label{eq:appendix_monotone_mj}
\end{equation}
because $\mathbf{S}\subseteq \mathbf{T}$. Since $a_j\ge 0$, summing over $j$ gives
\begin{equation}
F(\mathbf{S})=\sum_{j=1}^{T} a_j\, m_j(\mathbf{S})
\le
\sum_{j=1}^{T} a_j\, m_j(\mathbf{T})
= F(\mathbf{T}).
\label{eq:appendix_monotone_F}
\end{equation}
Hence $F$ is monotone non-decreasing.

\paragraph{(iii) Submodularity.}
Consider two subsets $\mathbf{S}\subseteq \mathbf{T}\subseteq \mathbf{V}$ and a candidate token $\mathbf{v}_u\in \mathbf{V}\setminus \mathbf{T}$. We first write the marginal gain of adding $\mathbf{v}_u$ to $\mathbf{S}$:
\begin{align}
\Delta(\mathbf{v}_u \mid \mathbf{S})
&=
F(\mathbf{S}\cup\{\mathbf{v}_u\}) - F(\mathbf{S}) \notag\\
&=
\sum_{j=1}^{T} a_j
\left(
m_j(\mathbf{S}\cup\{\mathbf{v}_u\}) - m_j(\mathbf{S})
\right).
\label{eq:appendix_submod_start}
\end{align}
By the definition of $m_j(\cdot)$,
\[
m_j(\mathbf{S}\cup\{\mathbf{v}_u\})
=
\max\!\left(m_j(\mathbf{S}),\, c_{uj}\right).
\]
Therefore,
\begin{align}
\Delta(\mathbf{v}_u \mid \mathbf{S})
&=
\sum_{j=1}^{T} a_j
\left(
\max\!\left(m_j(\mathbf{S}),\, c_{uj}\right)-m_j(\mathbf{S})
\right) \notag\\
&=
\sum_{j=1}^{T} a_j\, \max\!\left(c_{uj}-m_j(\mathbf{S}),\,0\right).
\label{eq:appendix_submod_marginal}
\end{align}

Now, since $\mathbf{S}\subseteq \mathbf{T}$, Eq.~\eqref{eq:appendix_monotone_mj} implies
\begin{equation}
m_j(\mathbf{S}) \le m_j(\mathbf{T}),\qquad \forall j.
\label{eq:appendix_submod_m_order}
\end{equation}
Fix any $j$. Because the function $x\mapsto \max(c_{uj}-x,0)$ is non-increasing in $x$, Eq.~\eqref{eq:appendix_submod_m_order} yields
\begin{equation}
\max\!\left(c_{uj}-m_j(\mathbf{S}),0\right)
\ge
\max\!\left(c_{uj}-m_j(\mathbf{T}),0\right).
\label{eq:appendix_submod_termwise}
\end{equation}
Multiplying both sides by $a_j\ge 0$ and summing over $j$, we obtain
\begin{equation}
\Delta(\mathbf{v}_u \mid \mathbf{S})
\ge
\Delta(\mathbf{v}_u \mid \mathbf{T}).
\label{eq:appendix_submod_final}
\end{equation}
This is exactly the diminishing-returns property. Hence $F$ is submodular.

Combining (i), (ii), and (iii), we conclude that $F$ is normalized, monotone, and submodular.
\hfill $\square$

\subsection{Approximation Guarantee of Greedy Optimization}

Since the ACPruner objective is NP-hard to optimize exactly, the natural question is whether the greedy algorithm used by ACPruner has a provable approximation guarantee. The answer is yes.

\paragraph{Corollary 1.}
Let $\mathbf{S}^{\star}$ be an optimal solution to
\begin{equation}
\max_{\mathbf{S}\subseteq \mathbf{V},\, |\mathbf{S}|\le k} F(\mathbf{S}),
\label{eq:appendix_opt_problem}
\end{equation}
and let $\mathbf{S}_{\mathrm{greedy}}$ be the subset returned by the greedy algorithm that iteratively selects the token with the largest marginal gain. Then
\begin{equation}
F(\mathbf{S}_{\mathrm{greedy}})
\ge
\left(1-\frac{1}{e}\right)F(\mathbf{S}^{\star}).
\label{eq:appendix_greedy_bound}
\end{equation}

\paragraph{Proof.}
Let $\mathbf{S}_t$ denote the greedy solution after $t$ selections, with $\mathbf{S}_0=\emptyset$, and let $\mathbf{v}_{g_t}$ be the token selected at step $t$. Then
\begin{equation}
\mathbf{S}_t = \mathbf{S}_{t-1}\cup\{\mathbf{v}_{g_t}\},
\qquad
\mathbf{v}_{g_t}
=
\arg\max_{\mathbf{v}\in \mathbf{V}\setminus \mathbf{S}_{t-1}}
\Delta(\mathbf{v}\mid \mathbf{S}_{t-1}).
\label{eq:appendix_greedy_rule}
\end{equation}

Fix any step $t\in\{1,\ldots,k\}$. Since $F$ is monotone,
\begin{equation}
F(\mathbf{S}^{\star}) \le F(\mathbf{S}_{t-1}\cup \mathbf{S}^{\star}).
\label{eq:appendix_opt_monotone}
\end{equation}
Let $\mathbf{S}^{\star}\setminus \mathbf{S}_{t-1}=\{\mathbf{v}_{u_1},\ldots,\mathbf{v}_{u_r}\}$, where $r\le k$. Then
\begin{align}
F(\mathbf{S}_{t-1}\cup \mathbf{S}^{\star}) - F(\mathbf{S}_{t-1})
&=
\sum_{s=1}^{r}
\Delta\!\left(
\mathbf{v}_{u_s}
\;\middle|\;
\mathbf{S}_{t-1}\cup \{\mathbf{v}_{u_1},\ldots,\mathbf{v}_{u_{s-1}}\}
\right)
\notag\\
&\le
\sum_{s=1}^{r}
\Delta(\mathbf{v}_{u_s}\mid \mathbf{S}_{t-1})
\label{eq:appendix_greedy_submod_use}
\end{align}
by submodularity. Since the greedy algorithm selects the largest marginal gain token,
\begin{equation}
\Delta(\mathbf{v}_{u_s}\mid \mathbf{S}_{t-1})
\le
\Delta(\mathbf{v}_{g_t}\mid \mathbf{S}_{t-1}),
\qquad \forall s.
\label{eq:appendix_greedy_best}
\end{equation}
Therefore,
\begin{equation}
F(\mathbf{S}_{t-1}\cup \mathbf{S}^{\star}) - F(\mathbf{S}_{t-1})
\le
r\, \Delta(\mathbf{v}_{g_t}\mid \mathbf{S}_{t-1})
\le
k\, \Delta(\mathbf{v}_{g_t}\mid \mathbf{S}_{t-1}).
\label{eq:appendix_greedy_gap_bound}
\end{equation}
Combining Eqs.~\eqref{eq:appendix_opt_monotone} and \eqref{eq:appendix_greedy_gap_bound}, we get
\begin{equation}
F(\mathbf{S}^{\star}) - F(\mathbf{S}_{t-1})
\le
k\, \Delta(\mathbf{v}_{g_t}\mid \mathbf{S}_{t-1}).
\label{eq:appendix_greedy_gap_bound2}
\end{equation}
Since
\[
\Delta(\mathbf{v}_{g_t}\mid \mathbf{S}_{t-1})
=
F(\mathbf{S}_t)-F(\mathbf{S}_{t-1}),
\]
Eq.~\eqref{eq:appendix_greedy_gap_bound2} becomes
\begin{equation}
F(\mathbf{S}_t)-F(\mathbf{S}_{t-1})
\ge
\frac{1}{k}\left(F(\mathbf{S}^{\star})-F(\mathbf{S}_{t-1})\right).
\label{eq:appendix_greedy_recurrence}
\end{equation}
Rearranging gives
\begin{equation}
F(\mathbf{S}^{\star})-F(\mathbf{S}_t)
\le
\left(1-\frac{1}{k}\right)
\left(F(\mathbf{S}^{\star})-F(\mathbf{S}_{t-1})\right).
\label{eq:appendix_greedy_recurrence2}
\end{equation}
Applying Eq.~\eqref{eq:appendix_greedy_recurrence2} recursively from $t=1$ to $t=k$, and using $F(\mathbf{S}_0)=F(\emptyset)=0$, we obtain
\begin{equation}
F(\mathbf{S}^{\star})-F(\mathbf{S}_k)
\le
\left(1-\frac{1}{k}\right)^k F(\mathbf{S}^{\star}).
\label{eq:appendix_greedy_recursive_final}
\end{equation}
Hence
\begin{equation}
F(\mathbf{S}_k)
\ge
\left(1-\left(1-\frac{1}{k}\right)^k\right)F(\mathbf{S}^{\star})
\ge
\left(1-\frac{1}{e}\right)F(\mathbf{S}^{\star}),
\label{eq:appendix_greedy_final}
\end{equation}
where the last inequality uses $\left(1-\frac{1}{k}\right)^k \le e^{-1}$. Since $\mathbf{S}_k=\mathbf{S}_{\mathrm{greedy}}$, the proof is complete.
\hfill $\square$

Corollary~1 shows that the greedy algorithm used by ACPruner is not merely a heuristic. Instead, it is a principled approximation algorithm for maximizing the proposed importance-weighted attention-induced coverage objective under a token budget.

\subsection{Generalized Coverage Transforms}

The ACPruner objective uses a linear utility on the coverage level of each token. We now show that the same theoretical structure still holds for a broader family of non-linear coverage utilities.

\paragraph{Theorem 2.}
For each token $\mathbf{v}_j$, let $\rho_j:[0,1]\rightarrow \mathbb{R}_{+}$ be a non-decreasing function satisfying $\rho_j(0)=0$. Define the generalized objective
\begin{equation}
F_{\rho}(\mathbf{S})
:=
\sum_{j=1}^{T} a_j\, \rho_j\!\left(\mathrm{Cov}(\mathbf{S}, \mathbf{v}_j)\right)
=
\sum_{j=1}^{T} a_j\, \rho_j\!\left(m_j(\mathbf{S})\right).
\label{eq:appendix_generalized_objective}
\end{equation}
Then $F_{\rho}$ is normalized, monotone, and submodular.

\paragraph{Proof.}
We again verify the three properties.

\paragraph{(i) Normalization.}
Since $\mathrm{Cov}(\emptyset,\mathbf{v}_j)=0$ and $\rho_j(0)=0$,
\begin{equation}
F_{\rho}(\emptyset)
=
\sum_{j=1}^{T} a_j\, \rho_j(0)
=
0.
\label{eq:appendix_generalized_normalized}
\end{equation}

\paragraph{(ii) Monotonicity.}
If $\mathbf{S}\subseteq \mathbf{T}$, then $m_j(\mathbf{S})\le m_j(\mathbf{T})$ for every $j$. Since each $\rho_j$ is non-decreasing,
\[
\rho_j(m_j(\mathbf{S})) \le \rho_j(m_j(\mathbf{T})).
\]
Multiplying by $a_j\ge 0$ and summing over $j$ yields
\begin{equation}
F_{\rho}(\mathbf{S}) \le F_{\rho}(\mathbf{T}).
\label{eq:appendix_generalized_monotone}
\end{equation}

\paragraph{(iii) Submodularity.}
Fix $\mathbf{S}\subseteq \mathbf{T}\subseteq \mathbf{V}$ and $\mathbf{v}_u\in \mathbf{V}\setminus \mathbf{T}$. For each token $j$, define the per-token marginal utility
\begin{equation}
\delta_j(\mathbf{v}_u \mid \mathbf{S})
:=
\rho_j\!\left(\max\{m_j(\mathbf{S}), c_{uj}\}\right)
-
\rho_j\!\left(m_j(\mathbf{S})\right).
\label{eq:appendix_delta_generalized}
\end{equation}
We will show
\begin{equation}
\delta_j(\mathbf{v}_u \mid \mathbf{S})
\ge
\delta_j(\mathbf{v}_u \mid \mathbf{T}),
\qquad \forall j.
\label{eq:appendix_delta_generalized_goal}
\end{equation}

Fix any $j$. Since $m_j(\mathbf{S})\le m_j(\mathbf{T})$, there are two cases.

\paragraph{Case 1: $c_{uj} \le m_j(\mathbf{T})$.}
Then
\[
\max\{m_j(\mathbf{T}),c_{uj}\}=m_j(\mathbf{T}),
\]
so
\[
\delta_j(\mathbf{v}_u \mid \mathbf{T})=0.
\]
On the other hand, because $\rho_j$ is non-decreasing,
\[
\delta_j(\mathbf{v}_u \mid \mathbf{S}) \ge 0.
\]
Hence Eq.~\eqref{eq:appendix_delta_generalized_goal} holds.

\paragraph{Case 2: $c_{uj} > m_j(\mathbf{T})$.}
Then necessarily $c_{uj}>m_j(\mathbf{S})$ as well, and thus
\begin{align}
\delta_j(\mathbf{v}_u \mid \mathbf{S})
&=
\rho_j(c_{uj})-\rho_j(m_j(\mathbf{S})), \\
\delta_j(\mathbf{v}_u \mid \mathbf{T})
&=
\rho_j(c_{uj})-\rho_j(m_j(\mathbf{T})).
\end{align}
Since $m_j(\mathbf{S})\le m_j(\mathbf{T})$ and $\rho_j$ is non-decreasing, we have
\[
\rho_j(m_j(\mathbf{S})) \le \rho_j(m_j(\mathbf{T})),
\]
which implies
\[
\delta_j(\mathbf{v}_u \mid \mathbf{S})
\ge
\delta_j(\mathbf{v}_u \mid \mathbf{T}).
\]

Therefore Eq.~\eqref{eq:appendix_delta_generalized_goal} holds in both cases. Multiplying by $a_j\ge 0$ and summing over $j$, we obtain
\begin{equation}
F_{\rho}(\mathbf{S}\cup\{\mathbf{v}_u\})-F_{\rho}(\mathbf{S})
\ge
F_{\rho}(\mathbf{T}\cup\{\mathbf{v}_u\})-F_{\rho}(\mathbf{T}),
\label{eq:appendix_generalized_submod}
\end{equation}
which proves submodularity.

Combining (i), (ii), and (iii), $F_{\rho}$ is normalized, monotone, and submodular.
\hfill $\square$

Note that Theorem~2 does not follow from a generic composition rule for submodular functions. Instead, it relies on the specific max-coverage structure of $m_j(\mathbf{S})$ in ACPruner.

Theorem~2 shows that the theoretical framework of ACPruner is not limited to the linear utility used in the main paper. More general monotone transforms of the coverage level preserve the same structural properties and thus remain amenable to greedy optimization.

\subsection{Lack of Uniform Guarantee for Importance-Only Selection}

We next show that token-wise importance alone is insufficient to guarantee a constant-factor approximation to the ACPruner objective.

\paragraph{Proposition 2.}
Let $\mathcal{A}$ be any deterministic selection rule that, given a token-wise score vector
\[
\mathbf{q}=(q_1,\ldots,q_T)
\]
and a budget $k$, outputs a subset $\mathcal{A}(\mathbf{q},k)\subseteq \mathbf{V}$ of size $k$. Suppose that $\mathcal{A}$ depends only on $\mathbf{q}$ and $k$, and is independent of the coverage matrix $(c_{ij})$. Then there does not exist any constant $\gamma>0$ such that
\begin{equation}
F\!\left(\mathcal{A}(\mathbf{q},k)\right)
\ge
\gamma \cdot
\max_{|\mathbf{S}|=k} F(\mathbf{S})
\label{eq:appendix_importance_uniform_goal}
\end{equation}
holds for all ACPruner instances.

\paragraph{Proof.}
We prove this by an explicit counterexample. Fix any deterministic rule $\mathcal{A}$ and any budget $k\ge 1$. Set the total number of tokens to
\[
T=k+1.
\]
Choose any score vector $\mathbf{q}$. Since $\mathcal{A}$ is deterministic, it outputs some subset
\[
\mathbf{S}_{\mathcal{A}} := \mathcal{A}(\mathbf{q},k), \qquad |\mathbf{S}_{\mathcal{A}}|=k.
\]
Because $T=k+1$, there exists at least one omitted token; let it be $\mathbf{v}_u\in \mathbf{V}\setminus \mathbf{S}_{\mathcal{A}}$.

Now construct an ACPruner instance as follows. First, define the importance weights by
\begin{equation}
a_u = 1,
\qquad
a_j = 0,\ \forall j\neq u.
\label{eq:appendix_importance_counter_weight}
\end{equation}
Next, define the coverage matrix as the identity matrix:
\begin{equation}
c_{ij}=
\begin{cases}
1, & \text{if } i=j,\\
0, & \text{if } i\neq j.
\end{cases}
\label{eq:appendix_importance_counter_cov}
\end{equation}
This is a valid ACPruner instance, since
\[
0\le c_{ij}\le 1,\qquad
\sum_{j=1}^{T} c_{ij}=1,\ \forall i,
\]
so the matrix satisfies the basic row-stochastic property inherited from attention weights.

Because $\mathbf{v}_u\notin \mathbf{S}_{\mathcal{A}}$, we have
\[
\max_{\mathbf{v}_i\in \mathbf{S}_{\mathcal{A}}} c_{iu}=0.
\]
Moreover, since $a_j=0$ for all $j\neq u$, the objective depends only on the coverage of token $\mathbf{v}_u$. Therefore,
\begin{equation}
F(\mathbf{S}_{\mathcal{A}})
=
\sum_{j=1}^{T} a_j \max_{\mathbf{v}_i\in \mathbf{S}_{\mathcal{A}}} c_{ij}
=
a_u \max_{\mathbf{v}_i\in \mathbf{S}_{\mathcal{A}}} c_{iu}
=
0.
\label{eq:appendix_importance_counter_alg}
\end{equation}

On the other hand, any subset $\mathbf{S}^{\star}$ of size $k$ that contains $\mathbf{v}_u$ satisfies
\[
\max_{\mathbf{v}_i\in \mathbf{S}^{\star}} c_{iu}=1,
\]
and hence
\begin{equation}
F(\mathbf{S}^{\star})
=
a_u \max_{\mathbf{v}_i\in \mathbf{S}^{\star}} c_{iu}
=
1.
\label{eq:appendix_importance_counter_opt}
\end{equation}
Therefore,
\[
\frac{F(\mathbf{S}_{\mathcal{A}})}{\max_{|\mathbf{S}|=k}F(\mathbf{S})}
=
0.
\]
Since the approximation ratio can be zero, no positive constant $\gamma$ can satisfy Eq.~\eqref{eq:appendix_importance_uniform_goal} uniformly over all instances. This proves the proposition.
\hfill $\square$

Proposition~2 formalizes the limitation of importance-only selection: a token may not stand out under a given token-wise ranking rule, yet still be indispensable because it is the unique representative of the only content that carries nonzero value under the ACPruner objective. This result is stated for deterministic selection rules, which already cover the standard score-ranking procedures used in practice.

\subsection{Lack of Uniform Guarantee for Diversity-Only Selection}

We finally show that internal diversity alone is also insufficient to guarantee a constant-factor approximation to the ACPruner objective.

\paragraph{Proposition 3.}
Let $\mathcal{B}$ be any deterministic selection rule that, given a pairwise dissimilarity matrix
\[
\mathbf{D}=(d_{ij})_{i,j=1}^{T}
\]
and a budget $k$, outputs a subset $\mathcal{B}(\mathbf{D},k)\subseteq \mathbf{V}$ of size $k$. Suppose that $\mathcal{B}$ depends only on $\mathbf{D}$ and $k$, and is independent of the importance weights $(a_j)$ and the coverage matrix $(c_{ij})$. Then there does not exist any constant $\gamma>0$ such that
\begin{equation}
F\!\left(\mathcal{B}(\mathbf{D},k)\right)
\ge
\gamma \cdot
\max_{|\mathbf{S}|=k} F(\mathbf{S})
\label{eq:appendix_diversity_uniform_goal}
\end{equation}
holds for all ACPruner instances.

\paragraph{Proof.}
Fix any deterministic rule $\mathcal{B}$ and any budget $k\ge 1$. Again set
\[
T=k+1.
\]
Choose any dissimilarity matrix $\mathbf{D}$. Since $\mathcal{B}$ is deterministic, it outputs some subset
\[
\mathbf{S}_{\mathcal{B}} := \mathcal{B}(\mathbf{D},k), \qquad |\mathbf{S}_{\mathcal{B}}|=k.
\]
Let $\mathbf{v}_u\in \mathbf{V}\setminus \mathbf{S}_{\mathcal{B}}$ be any omitted token.

Construct an ACPruner instance by defining
\begin{equation}
a_u = 1,
\qquad
a_j = 0,\ \forall j\neq u,
\label{eq:appendix_diversity_counter_weight}
\end{equation}
and again using the identity coverage matrix
\begin{equation}
c_{ij}=
\begin{cases}
1, & \text{if } i=j,\\
0, & \text{if } i\neq j.
\end{cases}
\label{eq:appendix_diversity_counter_cov}
\end{equation}
As above, this coverage matrix is a valid row-stochastic attention-induced instance.

Since $\mathbf{v}_u\notin \mathbf{S}_{\mathcal{B}}$, we have
\[
\max_{\mathbf{v}_i\in \mathbf{S}_{\mathcal{B}}} c_{iu}=0.
\]
Because all weights except $a_u$ are zero, it follows that
\begin{equation}
F(\mathbf{S}_{\mathcal{B}})
=
a_u \max_{\mathbf{v}_i\in \mathbf{S}_{\mathcal{B}}} c_{iu}
=
0.
\label{eq:appendix_diversity_counter_alg}
\end{equation}

However, any subset $\mathbf{S}^{\star}$ of size $k$ containing $\mathbf{v}_u$ satisfies
\[
\max_{\mathbf{v}_i\in \mathbf{S}^{\star}} c_{iu}=1,
\]
which gives
\begin{equation}
F(\mathbf{S}^{\star})
=
a_u \max_{\mathbf{v}_i\in \mathbf{S}^{\star}} c_{iu}
=
1.
\label{eq:appendix_diversity_counter_opt}
\end{equation}
Therefore,
\[
\frac{F(\mathbf{S}_{\mathcal{B}})}{\max_{|\mathbf{S}|=k}F(\mathbf{S})}
=
0.
\]
Hence no positive constant $\gamma$ can satisfy Eq.~\eqref{eq:appendix_diversity_uniform_goal} uniformly over all instances. This proves the proposition.
\hfill $\square$

Proposition~3 formalizes the limitation of diversity-only selection: even if a subset is highly dispersed internally, such dispersion alone does not guarantee that it covers the only content that matters under the ACPruner objective. Taken together with Proposition~2, these results theoretically justify why ACPruner must jointly model both content importance and attention-induced coverage.

\section{NP-hardness of the Optimization Problem}
\label{appendix:nphard}

In this section, we show that the optimization problem in Eq.~\eqref{eq:final_objective} is NP-hard in general. To establish this result, we consider its associated decision version.

\paragraph{Decision version.}
For a fixed $\beta > 0$, given nonnegative importance weights $\{w_j\}_{j=1}^{T}$, nonnegative coverage coefficients $\{c_{ij}\}_{i,j=1}^{T}$, a budget $k$, and a threshold $B$, the decision problem asks whether there exists a subset $\mathbf{S} \subseteq \mathbf{V}$ with $|\mathbf{S}| = k$ such that
\begin{equation}
\sum_{j=1}^{T} w_j^{\beta}\,\mathrm{Cov}(\mathbf{S}, \mathbf{v}_j)
=
\sum_{j=1}^{T} w_j^{\beta}\max_{\mathbf{v}_i \in \mathbf{S}} c_{ij}
\;\ge\; B.
\label{eq:nphard_decision}
\end{equation}
The threshold $B$ is introduced only for the decision version used in the complexity proof; it is not a variable in the main method.

\paragraph{Reduction from Maximum Coverage.}
We prove NP-hardness by a polynomial-time reduction from the classical \textsc{Maximum Coverage} problem. An instance of \textsc{Maximum Coverage} consists of
\begin{itemize}
    \item a universe $\mathcal{U} = \{e_1,\ldots,e_n\}$,
    \item a collection of subsets $\mathcal{A} = \{A_1,\ldots,A_m\}$ with $A_i \subseteq \mathcal{U}$,
    \item a budget $k$, and
    \item a threshold $B$.
\end{itemize}
The decision problem asks whether there exists an index set $I \subseteq \{1,\ldots,m\}$ with $|I| \le k$ such that
\begin{equation}
\left| \bigcup_{i \in I} A_i \right| \ge B.
\label{eq:maxcov_decision}
\end{equation}

We now construct an instance of Eq.~\eqref{eq:nphard_decision} from an arbitrary instance of Eq.~\eqref{eq:maxcov_decision}.

\paragraph{Construction.}
We create a token set
\begin{equation}
\mathbf{V} = \{\mathbf{s}_1,\ldots,\mathbf{s}_m,\mathbf{t}_1,\ldots,\mathbf{t}_n\},
\end{equation}
where each \textbf{selector token} $\mathbf{s}_i$ corresponds to subset $A_i$, and each \textbf{element token} $\mathbf{t}_j$ corresponds to universe element $e_j$. Hence the total number of tokens is $T = m+n$.

We define the importance weights as
\begin{equation}
w(\mathbf{s}_i) = 0,
\qquad
w(\mathbf{t}_j) = 1,
\label{eq:nphard_weights}
\end{equation}
for all $i \in \{1,\ldots,m\}$ and $j \in \{1,\ldots,n\}$. Since $\beta > 0$ is fixed, we have $1^\beta = 1$ and $0^\beta = 0$, so the exponent $\beta$ does not affect the reduction.

Next, we define the token-level coverage coefficients. For each selector token $\mathbf{s}_i$ and element token $\mathbf{t}_j$, let
\begin{equation}
c_{\mathbf{s}_i,\mathbf{t}_j}
=
\begin{cases}
1, & \text{if } e_j \in A_i,\\
0, & \text{otherwise}.
\end{cases}
\label{eq:nphard_selector_to_element}
\end{equation}
For all other pairs, we set
\begin{equation}
c_{ab} = 0.
\label{eq:nphard_other_pairs}
\end{equation}
Thus, only selector tokens can cover element tokens, and no other pair contributes to the objective.

\paragraph{Objective under the construction.}
For any subset $\mathbf{S} \subseteq \mathbf{V}$, the objective in Eq.~\eqref{eq:nphard_decision} becomes
\begin{equation}
\sum_{j=1}^{n} \max_{\mathbf{v} \in \mathbf{S}} c_{\mathbf{v},\mathbf{t}_j},
\label{eq:nphard_reduced_obj}
\end{equation}
because selector tokens have zero importance and all non-selector-to-element coverage terms are zero. Therefore, Eq.~\eqref{eq:nphard_reduced_obj} simply counts how many element tokens are covered by the selected selector tokens.

\paragraph{Correctness of the reduction.}
We now show that the \textsc{Maximum Coverage} instance is a \textsc{yes}-instance if and only if the constructed instance of Eq.~\eqref{eq:nphard_decision} is a \textsc{yes}-instance.

\textbf{(\(\Rightarrow\))} Suppose Eq.~\eqref{eq:maxcov_decision} is a \textsc{yes}-instance. Then there exists $I \subseteq \{1,\ldots,m\}$ with $|I| \le k$ such that
\[
\left| \bigcup_{i \in I} A_i \right| \ge B.
\]

Construct $\mathbf{S}$ by selecting the corresponding selector tokens $\{\mathbf{s}_i : i \in I\}$. If $|I| < k$, pad $\mathbf{S}$ with arbitrary additional tokens until $|\mathbf{S}|=k$. Such padding cannot decrease the objective value, since all coverage coefficients are nonnegative. By construction, for each element token $\mathbf{t}_j$,
\[
\max_{\mathbf{v} \in \mathbf{S}} c_{\mathbf{v},\mathbf{t}_j} = 1
\quad \Longleftrightarrow \quad
e_j \in \bigcup_{i \in I} A_i.
\]
Hence the objective value equals
\[
\left| \bigcup_{i \in I} A_i \right| \ge B.
\]
Therefore, the constructed instance of Eq.~\eqref{eq:nphard_decision} is also a \textsc{yes}-instance.

\textbf{(\(\Leftarrow\))} Conversely, suppose the constructed instance of Eq.~\eqref{eq:nphard_decision} is a \textsc{yes}-instance. Then there exists a subset $\mathbf{S} \subseteq \mathbf{V}$ with $|\mathbf{S}|=k$ such that
\[
\sum_{j=1}^{n} \max_{\mathbf{v} \in \mathbf{S}} c_{\mathbf{v},\mathbf{t}_j} \ge B.
\]
Let
\[
I = \{\, i \in \{1,\ldots,m\} \mid \mathbf{s}_i \in \mathbf{S} \,\}
\]
be the indices of selector tokens chosen in $\mathbf{S}$. Since only selector tokens can cover element tokens, the objective value is exactly
\[
\left| \bigcup_{i \in I} A_i \right|.
\]
Moreover, $|I| \le |\mathbf{S}| = k$. Therefore,
\[
\left| \bigcup_{i \in I} A_i \right| \ge B,
\]
which means that the original \textsc{Maximum Coverage} instance is also a \textsc{yes}-instance.

\paragraph{Conclusion.}
The above reduction is polynomial-time, so the decision version of Eq.~\eqref{eq:final_objective} is NP-hard. Consequently, the optimization problem in Eq.~\eqref{eq:final_objective} is also NP-hard, since an exact polynomial-time algorithm for the optimization problem would immediately solve its decision version.

\section{Pseudocode of ACPruner}
\label{appendix:pseudocode}

Algorithm~\ref{alg:acpruner} summarizes the complete pipeline of ACPruner. Given the visual token set, the textual query, and the vision encoder attention maps, ACPruner first computes the fused importance weights for all original visual tokens, then derives the attention-induced token-to-token coverage matrix, and finally performs greedy subset selection under a fixed token budget.

\begin{algorithm}[H]
\caption{ACPruner}
\label{alg:acpruner}
\KwIn{
visual tokens $\mathbf{V} = [\mathbf{v}_1,\ldots,\mathbf{v}_T]$; \\
text tokens $\mathbf{U} = [\mathbf{u}_1,\ldots,\mathbf{u}_N]$; \\
vision encoder attentions $\{A^{(l,h)}\}$; \\
budget $k$; trade-off coefficient $\alpha$; bias exponent $\beta$
}
\KwOut{selected token subset $\mathbf{S}$}

\BlankLine
\textbf{// Step 1: Compute intra-modality importance} \\
\If{the vision encoder contains a CLS token}{
Compute $s_j^{\mathrm{intra}}$ using penultimate-layer CLS-to-patch attention by Eq.~\eqref{eq:intra_cls} \;
}
\Else{
Compute $s_j^{\mathrm{intra}}$ using average incoming patch attention \;
}

\BlankLine
\textbf{// Step 2: Compute inter-modality importance} \\
Extract noun tokens $\mathcal{N}(\mathbf{U})$ from the input text \;
\If{$\mathcal{N}(\mathbf{U}) = \emptyset$}{
Set $\alpha \leftarrow 1$ \;
\For{$j = 1$ to $T$}{
Set $s_j^{\mathrm{inter}} \leftarrow 0$ \;
}
}
\Else{
\For{$j = 1$ to $T$}{
Compute $s_j^{\mathrm{inter}} = \max_{\mathbf{u}_k \in \mathcal{N}(\mathbf{U})} r_{jk}$ by Eq.~\eqref{eq:inter_score} \;
}
}

\BlankLine
\textbf{// Step 3: Fuse token importance} \\
Normalize $\{s_j^{\mathrm{intra}}\}_{j=1}^{T}$ and $\{s_j^{\mathrm{inter}}\}_{j=1}^{T}$ by softmax to obtain
$\{\tilde{s}_j^{\mathrm{intra}}\}_{j=1}^{T}$ and $\{\tilde{s}_j^{\mathrm{inter}}\}_{j=1}^{T}$ \;
\For{$j = 1$ to $T$}{
Compute fused importance weight
$w_j = \alpha \tilde{s}_j^{\mathrm{intra}} + (1-\alpha)\tilde{s}_j^{\mathrm{inter}}$ by Eq.~\eqref{eq:fused_importance} \;
Compute biased weight $\bar{w}_j = w_j^{\beta}$ \;
}

\BlankLine
\textbf{// Step 4: Compute attention-induced token coverage} \\
\For{$i = 1$ to $T$}{
    \For{$j = 1$ to $T$}{
        Compute token-level coverage
        $c_{ij} = \frac{1}{LH}\sum_{l=1}^{L}\sum_{h=1}^{H} A^{(l,h)}_{i \rightarrow j}$
        by Eq.~\eqref{eq:token_coverage} \;
    }
}

\BlankLine
\textbf{// Step 5: Greedy subset selection} \\
Initialize $\mathbf{S} \leftarrow \emptyset$ \;
Initialize $m_j \leftarrow 0$ for all $j \in \{1,\ldots,T\}$ \;

\While{$|\mathbf{S}| < k$}{
    \ForEach{$\mathbf{v}_i \in \mathbf{V} \setminus \mathbf{S}$}{
        Compute marginal gain
        \[
        \Delta(\mathbf{v}_i \mid \mathbf{S})
        =
        \sum_{j=1}^{T}
        \bar{w}_j \,
        \max(c_{ij} - m_j,\, 0)
        \]
        according to Eq.~\eqref{eq:marginal_gain} \;
    }
    Select
    \[
    \mathbf{v}_{i^\star}
    =
    \arg\max_{\mathbf{v}_i \in \mathbf{V} \setminus \mathbf{S}}
    \Delta(\mathbf{v}_i \mid \mathbf{S})
    \]
    \;
    Update $\mathbf{S} \leftarrow \mathbf{S} \cup \{\mathbf{v}_{i^\star}\}$ \;
    \For{$j = 1$ to $T$}{
        Update current coverage
        $m_j \leftarrow \max(m_j, c_{i^\star j})$ \;
    }
}

\Return{$\mathbf{S}$}
\end{algorithm}

\section{Results on Larger Model Scales}
\label{results-13b}

\begin{table}[H]
\caption{Performance comparison on LLaVA-1.5-13B under different pruning ratios. ``Perf. Ret.'' denotes the average performance retention rate relative to the full-token baseline.}
\label{tab:llava-1.5-13b}
\resizebox{\columnwidth}{!}{
\begin{tabular}{l|c|ccccccccc|c}
\toprule
\rowcolor[HTML]{F2F3F5} 
\textbf{Method}   & \textbf{Venue} & \textbf{MME}  & \textbf{GQA}  & \textbf{SQA} & \textbf{POPE} & \textbf{TextVQA} & \textbf{VizWiz} & \textbf{VQA-v2} & \textbf{MMB} & \textbf{MM-Vet} & \textbf{Perf. Ret.} \\
\midrule
\rowcolor[HTML]{F5F6F7} 
\multicolumn{12}{c}{\cellcolor[HTML]{F5F6F7}  Upper Bound, 576 visual tokens  (pruning ratio = 0\%)}                                                                                                 \\
\midrule
LLaVA-1.5-13B     & CVPR'24        & 1818          & 63.2          & 72.8           & 85.9          & 61.3             & 56.6            & 80.0            & 67.7            & 35.3            & 100.0\%             \\
\midrule
\rowcolor[HTML]{F5F6F7} 
\multicolumn{12}{c}{\cellcolor[HTML]{F5F6F7}  Retain 192 visual tokens (pruning ratio = 66.7\%)}                                                                                                     \\
\midrule
VisionZIP         & CVPR'25        & 1754          & 59.1          & 73.5           & 85.1          & 59.5             & 54.9            & 78.0            & 66.9            & 37.5            & 98.5\%              \\
SCOPE             & NeurIPS'25     & 1775          & 59.7          & \textbf{73.8}  & 86.7          & 60.0             & 55.0            & 78.1            & \textbf{67.6}   & \textbf{39.4}   & 99.8\%              \\
PruneSID          & ICLR'26        & 1770          & 59.6          & 72.8           & 86.4          & 58.6             & 56.0            & 78.0            & 65.9            & 38.0            & 98.8\%              \\

\rowcolor[HTML]{F0F4FF} 
\textbf{ACPruner} & \textbf{Ours}  & \textbf{1878} & \textbf{60.3} & 73.6           & \textbf{86.9} & \textbf{60.4}    & \textbf{56.2}   & \textbf{78.8}   & 67.2            & 38.4            & \textbf{100.6\%}    \\
\midrule
\rowcolor[HTML]{F5F6F7} 
\multicolumn{12}{c}{\cellcolor[HTML]{F5F6F7}  Retain 128 visual tokens (pruning ratio = 77.8\%)}                                                                                                     \\
\midrule
VisionZIP         & CVPR'25        & 1743          & 57.9          & 74.0           & 85.2          & 58.7             & 55.0            & 76.8            & 66.7            & 37.5            & 98.0\%              \\
SCOPE             & NeurIPS'25     & 1735          & 59.3          & 73.9           & 85.9          & 58.7             & 56.0            & 77.0            & 67.2            & \textbf{37.7}   & 98.6\%              \\
PruneSID          & ICLR'26        & 1811          & 58.9          & 73.1           & 85.9          & 57.5             & \textbf{56.8}   & 76.7            & 65.5            & 37.3            & 98.4\%              \\
\rowcolor[HTML]{F0F4FF} 
\textbf{ACPruner} & \textbf{Ours}  & \textbf{1829} & \textbf{59.6} & \textbf{74.5}  & \textbf{86.1} & \textbf{59.1}    & 56.2            & \textbf{77.4}   & \textbf{67.2}   & 37.5            & \textbf{99.4\%}     \\
\midrule
\rowcolor[HTML]{F5F6F7} 
\multicolumn{12}{c}{\cellcolor[HTML]{F5F6F7}  Retain 64 visual tokens (pruning ratio = 88.9\%)}                                                                                                      \\
\midrule
VisionZIP         & CVPR'25        & 1676          & 56.2          & \textbf{74.4}  & 76.0          & 57.4             & 55.9            & 73.7            & 64.9            & 33.9            & 94.2\%              \\
SCOPE             & NeurIPS'25     & 1762          & 58.7          & 73.2           & 83.0          & 58.3             & 56.8            & 75.0            & 65.5            & 35.7            & 97.3\%              \\
PruneSID          & ICLR'26        & 1711          & 57.8          & 71.8           & 82.0          & 56.3             & 57.3            & 75.2            & 63.8            & 35.2            & 95.6\%              \\
\rowcolor[HTML]{F0F4FF} 
\textbf{ACPruner} & \textbf{Ours}  & \textbf{1820} & \textbf{58.7} & 72.7           & \textbf{85.4} & \textbf{58.5}    & \textbf{57.5}   & \textbf{75.8}   & \textbf{66.2}   & \textbf{35.8}   & \textbf{98.1\%}    \\
\bottomrule
\end{tabular}}
\end{table}

\begin{table}[H]
\caption{Performance comparison on LLaVA-NeXT-13B under different pruning ratios.
``Perf. Ret.'' denotes the average performance retention rate relative to the full-token baseline.}
\label{tab:llava-next-13b}
\resizebox{\columnwidth}{!}{
\begin{tabular}{l|c|cccccccc|c}
\toprule
\rowcolor[HTML]{F2F3F5} 
\textbf{Method}   & \textbf{Venue} & \textbf{MME}  & \textbf{GQA}  & \textbf{SQA} & \textbf{POPE} & \textbf{TextVQA} & \textbf{VizWiz} & \textbf{VQA-v2} & \textbf{MMB} & \textbf{Perf. Ret.} \\
\midrule
\rowcolor[HTML]{F5F6F7} 
\multicolumn{11}{c}{\cellcolor[HTML]{F5F6F7} Upper Bound, 2880 visual tokens  (pruning ratio = 0\%)}                                                                              \\
\midrule
LLaVA-NeXT-13B    & CVPR'24        & 1901          & 65.4          & 73.5           & 86.2          & 64.3             & 64.0            & 81.8            & 70.0            & 100.0\%             \\
\midrule
\rowcolor[HTML]{F5F6F7} 
\multicolumn{11}{c}{\cellcolor[HTML]{F5F6F7} Retain up to 640 visual tokens (pruning ratio = 77.8\%)}                                                                             \\
\midrule
VisionZIP         & CVPR'25        & 1871          & 63.0          & 71.2           & 85.7          & 62.2             & 56.0            & 79.7            & 68.6            & 96.3\%              \\
SCOPE             & NeurIPS'25     & \textbf{1897} & 63.6          & 72.5           & 86.4          & 62.4             & 60.0            & 79.5            & 69.3            & 97.9\%              \\
PruneSID          & ICLR'26        & 1817          & 62.4          & 70.1           & 85.6          & 60.2             & 60.2            & 79.1            & 67.0            & 96.1\%              \\
\rowcolor[HTML]{F0F4FF} 
\textbf{ACPruner} & \textbf{Ours}  & 1894          & \textbf{64.3} & \textbf{73.7}  & \textbf{86.5} & \textbf{63.0}    & \textbf{60.8}   & \textbf{80.1}   & \textbf{70.2}   & \textbf{98.7\%}     \\
\midrule
\rowcolor[HTML]{F5F6F7} 
\multicolumn{11}{c}{\cellcolor[HTML]{F5F6F7} Retain up to 320 visual tokens (pruning ratio = 88.9\%)}                                                                             \\
\midrule
VisionZIP         & CVPR'25        & 1805          & 60.7          & 70.3           & 82.0          & 60.9             & 56.7            & 76.8            & 67.2            & 93.9\%              \\
SCOPE             & NeurIPS'25     & 1830          & 63.0          & 71.7           & 85.1          & 60.8             & 57.7            & 77.3            & 67.7            & 95.6\%              \\
PruneSID          & ICLR'26        & 1810          & 61.5          & 70.3           & 82.7          & 58.5             & \textbf{58.5}   & 76.9            & 65.4            & 93.8\%              \\
\rowcolor[HTML]{F0F4FF} 
\textbf{ACPruner} & \textbf{Ours}  & \textbf{1848} & \textbf{63.5} & \textbf{73.4}  & \textbf{85.7} & \textbf{61.6}    & 58.4            & \textbf{77.8}   & \textbf{68.2}   & \textbf{96.6\%}     \\
\midrule
\rowcolor[HTML]{F5F6F7} 
\multicolumn{11}{c}{\cellcolor[HTML]{F5F6F7} Retain up to 160 visual tokens (pruning ratio = 94.4\%)}                                                                             \\
\midrule
VisionZIP         & CVPR'25        & 1739          & 57.8          & 69.3           & 76.6          & 58.4             & 55.3            & 72.4            & 64.9            & 90.2\%              \\
SCOPE             & NeurIPS'25     & 1777          & 61.4          & 72.0           & 82.8          & 59.3             & 56.3            & 74.0            & \textbf{66.9}   & 93.5\%              \\
PruneSID          & ICLR'26        & 1715          & 59.5          & 69.1           & 77.8          & 56.7             & 56.7            & 73.6            & 65.5            & 90.7\%              \\
\rowcolor[HTML]{F0F4FF} 
\textbf{ACPruner} & \textbf{Ours}  & \textbf{1797} & \textbf{62.2} & \textbf{72.9}  & \textbf{84.9} & \textbf{60.1}    & \textbf{57.0}   & \textbf{74.9}   & 66.0            & \textbf{94.5\%}    \\
\bottomrule
\end{tabular}
}
\end{table}

To further evaluate the scalability of ACPruner, we extend our experiments to larger 13B-scale LVLMs, including LLaVA-1.5-13B and LLaVA-NeXT-13B. The results are respectively reported in Tables~\ref{tab:llava-1.5-13b} and~\ref{tab:llava-next-13b}.

\paragraph{Results on LLaVA-1.5-13B.}
Table~\ref{tab:llava-1.5-13b} shows that ACPruner consistently achieves the best overall performance retention across all three pruning ratios. At the pruning ratio of 66.7\%, ACPruner reaches 100.6\% performance retention, slightly surpassing the full-token baseline. At the more aggressive pruning ratio of 77.8\%, it still delivers the best average performance retention of 99.4\%. Even under the extremely high pruning ratio of 88.9\%, ACPruner remains the strongest method overall, achieving 98.1\% performance retention. These results are consistent with those on LLaVA-1.5-7B, showing that the advantages of ACPruner transfer well to larger model scales.

\paragraph{Results on LLaVA-NeXT-13B.}
Table~\ref{tab:llava-next-13b} reports the results on LLaVA-NeXT-13B. ACPruner again achieves the best overall performance retention under all pruning ratios. At the pruning ratio of 77.8\%, it obtains 98.7\% performance retention, outperforming all comparison methods. At 88.9\% pruning, ACPruner still delivers the best overall performance retention of 96.6\%. Even under the extremely high pruning ratio of 94.4\%, where only up to 160 visual tokens are retained, ACPruner remains the best-performing method with 94.5\% performance retention. These results further verify the robustness of ACPruner on larger high-resolution LVLMs.

\section{Additional Ablation Study}
\label{appendix-ablation}
In this section, we present additional ablation results to further analyze the impact of various design choices in ACPruner.

\subsection{Ablation of Intra-modality Importance Variants}

In ACPruner, the intra-modality importance of a visual token is computed to capture its contribution to the global image semantics, typically using CLS-to-patch attention in the penultimate layer of the vision encoder (or the average incoming attention if no CLS token exists). To further evaluate the design choices for this component, we compare four variants on LLaVA-1.5-7B under a 77.8\% pruning ratio: CLS attention, patch incoming attention, L2 norm, and similarity with the global mean. CLS attention uses the attention from the CLS token to each patch token in the penultimate encoder layer; patch incoming attention uses the average incoming attention received by each patch token across all encoder layers; L2 norm uses the feature magnitude of each token; and similarity with the global mean measures the alignment between each token and the global mean visual feature.

Table~\ref{tab:ablation-intra} shows that CLS attention achieves the best overall performance, with the highest performance retention of 98.9\%. Replacing it with patch incoming attention causes a small but consistent drop, indicating that it is a reasonable surrogate but weaker than CLS attention as a global saliency cue. The two feature-based variants perform noticeably worse: L2 norm leads to a clear degradation, while similarity with global mean performs worst overall, especially on TextVQA. These results show that attention-based global importance is more effective than simple feature magnitude or feature similarity for identifying visually important tokens, justifying our choice of CLS attention as the default intra-modality importance when a CLS token is available.

\begin{table}[H]
\centering
\caption{Ablation of different intra-modality importance variants on LLaVA-1.5-7B under a 77.8\% pruning ratio. The blue row corresponds to ACPruner's design choice.}
\label{tab:ablation-intra}
\begin{tabular}{l|ccccc|c}
\toprule
\rowcolor[HTML]{F2F3F5} 
\textbf{}                   & \textbf{MME} & \textbf{GQA} & \textbf{SQA} & \textbf{POPE} & \textbf{TextVQA} & \textbf{Perf. Ret.} \\
\midrule
\rowcolor[HTML]{F0F4FF} 
CLS attention               & 1812         & 59.8         & 69.4         & 87.2          & 57.6             & 98.9\%              \\
Patch incoming attention     & 1780         & 59.2         & 68.9         & 87.1          & 57.1             & 98.0\%              \\
L2 norm                     & 1707         & 59.7         & 68.5         & 87.0          & 53.9             & 96.1\%              \\
Similarity with global mean & 1665         & 59.1         & 68.6         & 87.2          & 48.8             & 93.8\%           
\\
\bottomrule
\end{tabular}
\end{table}

\subsection{Ablation of Noun Extraction}

In ACPruner, noun extraction is used to identify the key semantic units of the input query, providing the textual cues that guide which visual tokens should be prioritized. To evaluate the impact of this design choice, we ablate different noun extraction strategies on LLaVA-1.5-7B under a 77.8\% pruning ratio. We compare four settings: using only nouns, using nouns with adjectives, using nouns with adjectives and verbs, and using the full text. Table~\ref{tab:ablation-noun} shows that using only nouns achieves the best overall performance. As more linguistic categories are included, the performance gradually declines, and using the full text performs worst. These results suggest that nouns provide the cleanest semantic cues for identifying query-relevant visual content, while introducing additional modifiers or the full text tends to bring more irrelevant linguistic noise. This justifies our choice of using noun extraction as the default text processing strategy.

\begin{table}[H]
\centering
\caption{Ablation of noun extraction on LLaVA-1.5-7B under a 77.8\% pruning ratio. The blue row corresponds to ACPruner's design choice.}
\label{tab:ablation-noun}
\begin{tabular}{l|cccc|c}
\toprule
\rowcolor[HTML]{F2F3F5} 
\textbf{}          & \textbf{MME} & \textbf{GQA} & \textbf{SQA} & \textbf{POPE} & \textbf{Perf. Ret.} \\
\midrule
\rowcolor[HTML]{F0F4FF} 
Noun              & 1812         & 59.8         & 69.4         & 87.2          & 98.8\%              \\
Noun + Adj        & 1804         & 59.8         & 69.2         & 87.2          & 98.6\%              \\
Noun + Adj + Verb & 1800         & 59.7         & 69.1         & 87.2          & 98.5\%              \\
Full text          & 1782         & 59.5         & 68.9         & 86.9          & 98.0\%             \\
\bottomrule
\end{tabular}
\end{table}

\subsection{Ablation of Inter-modality Importance Variants}
In ACPruner, inter-modality importance is used to measure how relevant each visual token is to the input textual query. To further evaluate the design choices for this component, we ablate different variants on LLaVA-1.5-7B under a 77.8\% pruning ratio. We compare two design choices: the matching granularity, including token-level and word-level matching, and the aggregation strategy, including max and mean pooling. Table~\ref{tab:ablation-inter} shows that Word-level Max achieves the best overall performance. Compared with token-level matching, word-level matching consistently performs better, indicating that aggregating subword tokens into word-level semantic units provides a cleaner relevance signal. Moreover, max pooling consistently outperforms mean pooling under both token-level and word-level settings, suggesting that visual relevance is better characterized by the strongest alignment to a key textual unit rather than by average similarity over all textual units. These results justify our choice of word-level max similarity as the default definition of inter-modality importance.

\begin{table}[H]
\centering
\caption{Ablation of different inter-modality importance variants on LLaVA-1.5-7B under a 77.8\% pruning ratio. The blue row corresponds to ACPruner's design choice.}
\label{tab:ablation-inter}
\begin{tabular}{l|ccccc|c}
\toprule
\rowcolor[HTML]{F2F3F5} 
\textbf{}        & \textbf{MME} & \textbf{GQA} & \textbf{SQA} & \textbf{POPE} & \textbf{TextVQA} & \textbf{Perf. Ret.} \\
\midrule
Token-level Max  & 1813         & 59.7         & 69.2         & 86.9          & 57.5             & 98.7\%              \\
Token-level Mean & 1806         & 59.6         & 68.7         & 86.8          & 56.9             & 98.2\%              \\
\rowcolor[HTML]{F0F4FF} 
Word-level Max   & 1812         & 59.8         & 69.4         & 87.2          & 57.6             & 98.9\%              \\
Word-level Mean  & 1768         & 59.3         & 69.2         & 86.0          & 57.4             & 97.8\%      \\
\bottomrule
\end{tabular}
\end{table}

\subsection{Ablation of Encoder Attention Layer Choices}

ACPruner uses encoder attention in two places. First, for intra-modality importance, it uses the CLS-to-patch attention from the penultimate layer. Second, for token coverage, it uses the patch-to-patch attention averaged across all layers. Since prior studies have shown that attention patterns in ViT-based vision encoders vary substantially across different depths~\cite{hero,hired}, we further ablate the choice of attention layers for these two components.

Table~\ref{tab:ablation-cls-layer} reports the ablation results of the CLS attention layer used for intra-modality importance. The best performance is achieved by using the penultimate layer (layer 22), which matches our default design. Earlier layers and middle layers perform consistently worse, and averaging CLS attention across all layers also leads to a clear drop. This suggests that the final-stage CLS attention provides the strongest global saliency signal for identifying visually important tokens.

Table~\ref{tab:ablation-patch-layer} reports the ablation results of the patch attention layer used for token coverage. In contrast to CLS attention, the best performance is obtained by averaging patch-to-patch attention across all layers, which again matches our default design. Using only a single layer, whether early, middle, or late, consistently underperforms the all-layer setting. This indicates that token-to-token coverage is better characterized by aggregating attention information across the full encoding process, rather than relying on any individual layer alone.

\begin{table}[H]
\centering
\caption{Ablation of CLS attention layer on LLaVA-1.5-7B under a 77.8\% pruning ratio. The blue row corresponds to ACPruner's design choice.}
\label{tab:ablation-cls-layer}
\begin{tabular}{l|ccccc|c}
\toprule
\rowcolor[HTML]{F2F3F5} 
\textbf{Layer Index} & \textbf{MME} & \textbf{GQA} & \textbf{SQA} & \textbf{POPE} & \textbf{TextVQA} & \textbf{Perf. Ret.} \\
\midrule
\rowcolor[HTML]{F5F6F7} 
\multicolumn{7}{c}{\cellcolor[HTML]{F5F6F7} Initial Layers}                                               \\
\midrule
0                    & 1797         & 59.5         & 68.8         & 87.1          & 55.1             & 97.5\%              \\
1                    & 1746         & 59.7         & 69.2         & 87.2          & 53.6             & 96.7\%              \\
\midrule
\rowcolor[HTML]{F5F6F7} 
\multicolumn{7}{c}{\cellcolor[HTML]{F5F6F7} Middle Layers}                                                \\
\midrule
9                    & 1753         & 59.4         & 68.3         & 87.0          & 55.6             & 97.0\%              \\
10                   & 1756         & 59.5         & 68.1         & 86.8          & 55.7             & 97.0\%              \\
\midrule
\rowcolor[HTML]{F5F6F7} 
\multicolumn{7}{c}{\cellcolor[HTML]{F5F6F7} Final Layers}                                                 \\
\midrule
\rowcolor[HTML]{F0F4FF} 
22                   & 1812         & 59.8         & 69.4         & 87.2          & 57.6             & 98.9\%              \\
23                   & 1788         & 59.4         & 69.4         & 86.9          & 57.6             & 98.4\%                \\
\midrule
\rowcolor[HTML]{F5F6F7} 
\multicolumn{7}{c}{\cellcolor[HTML]{F5F6F7} All Layers}                                                   \\
\midrule
0-23                 & 1751         & 59.4         & 68.7         & 86.8          & 57.0             & 97.6\%       \\
\bottomrule
\end{tabular}
\end{table}

\begin{table}[H]
\centering
\caption{Ablation of patch attention layer on LLaVA-1.5-7B under a 77.8\% pruning ratio. The blue row corresponds to ACPruner's design choice.}
\label{tab:ablation-patch-layer}
\begin{tabular}{l|ccccc|c}
\toprule
\rowcolor[HTML]{F2F3F5} 
\textbf{Layer Index} & \textbf{MME} & \textbf{GQA} & \textbf{SQA} & \textbf{POPE} & \textbf{TextVQA} & \textbf{Perf. Ret.} \\
\midrule
\rowcolor[HTML]{F5F6F7} 
\multicolumn{7}{c}{\cellcolor[HTML]{F5F6F7} Initial Layers}                                               \\
\midrule
0                    & 1738         & 58.8         & 68.5         & 85.2          & 52.2             & 89.7\%              \\
1                    & 1745         & 59.4         & 68.4         & 86.3          & 53.7             & 92.3\%              \\
\midrule
\rowcolor[HTML]{F5F6F7} 
\multicolumn{7}{c}{\cellcolor[HTML]{F5F6F7} Middle Layers}                                                \\
\midrule
9                    & 1721         & 60.0         & 68.4         & 86.5          & 53.3             & 91.6\%              \\
10                   & 1784         & 59.5         & 68.1         & 86.6          & 56.3             & 96.7\%              \\
\rowcolor[HTML]{F5F6F7} 
\multicolumn{7}{c}{\cellcolor[HTML]{F5F6F7} Final Layers}                                                 \\
\midrule
22                   & 1746         & 59.6         & 68.5         & 87.0          & 55.0             & 94.5\%              \\
23                   & 1771         & 59.4         & 68.9         & 87.2          & 54.2             & 93.1\%              \\
\midrule
\rowcolor[HTML]{F5F6F7} 
\multicolumn{7}{c}{\cellcolor[HTML]{F5F6F7} All Layers}                                                   \\
\midrule
\rowcolor[HTML]{F0F4FF} 
0-23                 & 1812         & 59.8         & 69.4         & 87.2          & 57.6             & 98.9\%       \\
\bottomrule
\end{tabular}
\end{table}

\section{Step-wise Runtime Breakdown of ACPruner}
\label{appendix:runtime}

\begin{figure}[h]
    \centering
    \includegraphics[width=\columnwidth]{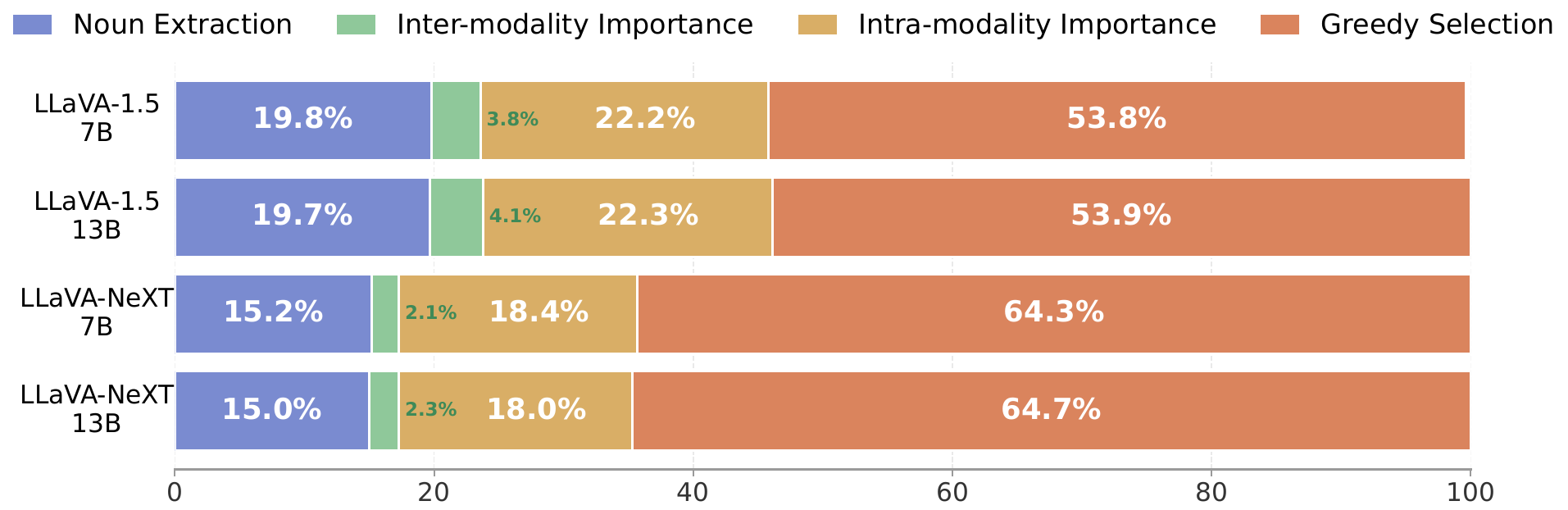}
    \caption{Step-wise runtime distribution of ACPruner on different LVLM backbones.}
    \label{fig:distribution}
\end{figure}
To further understand the practical overhead of ACPruner, we visualize the runtime distribution of its internal steps on different LVLM backbones in Fig.~\ref{fig:distribution}. We decompose the pruning pipeline into four parts: noun extraction, inter-modality importance computation, intra-modality importance computation, and greedy selection.
As shown in Fig.~\ref{fig:distribution}, greedy selection is the dominant cost within ACPruner across all backbones, accounting for more than half of the total pruning time in all cases. This proportion is larger on the LLaVA-NeXT family, where the initial number of visual tokens is higher. In contrast, inter-modality importance contributes the smallest runtime share, indicating that the text-guided relevance estimation introduces only a minor overhead. Noun extraction and intra-modality importance take a moderate portion of the total time, and their proportions remain relatively stable within each model family.

\section{Details of Evaluation Benchmarks}
\label{appendix:benchmarks}

We evaluate ACPruner on a diverse set of image and video benchmarks covering different multimodal tasks. Below, we briefly introduce the benchmarks used in our experiments.

\subsection{Image Benchmarks}

\noindent\textbf{MME~\cite{mme}.} MME is a comprehensive evaluation benchmark for multimodal large language models. It measures both perception and cognition abilities through 14 subtasks, and its instruction-answer pairs are manually designed to reduce data leakage and minimize the influence of prompt engineering. It is widely used to assess the general multimodal understanding capabilities of MLLMs in a controlled and holistic manner.

\noindent\textbf{GQA~\cite{gqa}.} GQA is a benchmark for real-world visual reasoning and compositional question answering. It leverages scene graph structures to generate 22M diverse reasoning questions, each paired with a functional program representing its semantics. In addition to standard accuracy, GQA emphasizes deeper evaluation dimensions such as consistency, grounding, and plausibility, making it well suited for assessing fine-grained reasoning over complex scenes.

\noindent\textbf{SQA / ScienceQA~\cite{sqa}.} ScienceQA is a multimodal science question answering benchmark consisting of approximately 21K multiple-choice questions covering a broad range of science topics. Beyond answer labels, it also provides corresponding lectures and explanations, enabling the evaluation of multimodal reasoning and knowledge-intensive problem solving. It is particularly useful for measuring a model's ability to integrate visual evidence with scientific commonsense and structured reasoning.

\noindent\textbf{POPE~\cite{pope}.} POPE is a benchmark designed to evaluate object hallucination in large vision-language models. It was introduced as part of the first systematic study of hallucination in LVLMs and uses a polling-based query method to provide a more stable and flexible evaluation protocol. POPE specifically tests whether a model tends to mention objects that are inconsistent with the visual content, making it a standard benchmark for hallucination analysis.

\noindent\textbf{TextVQA~\cite{textvqa}.} TextVQA focuses on question answering that requires models to read and reason about scene text in images. The dataset contains 45,336 questions on 28,408 images, where accurate answering depends not only on visual understanding but also on recognizing and interpreting embedded textual information. It is a representative benchmark for evaluating OCR-aware multimodal reasoning.

\noindent\textbf{VizWiz~\cite{vizwiz}.} VizWiz is a goal-oriented VQA benchmark built from real visual questions asked by blind people. It contains over 31,000 visual questions, where each image is captured by a blind photographer using a mobile phone and each question is recorded in spoken form. Compared with conventional VQA benchmarks, VizWiz is more challenging because the images are often of poor quality, the questions are more conversational, and many questions are unanswerable from the image alone.

\noindent\textbf{VQA-v2~\cite{vqav2}.} VQA-v2 is a balanced visual question answering benchmark designed to make the visual content matter more. It extends the original VQA dataset by collecting complementary image pairs such that the same question can correspond to different answers on similar images, thereby reducing the effect of language priors. This benchmark is widely used to evaluate open-ended visual understanding and question answering on real-world images.

\noindent\textbf{MMB-en~\cite{mmb}.} MMBench is a systematically designed objective benchmark for the holistic evaluation of vision-language models, and it provides both English and Chinese multiple-choice questions; in our experiments, we use its English subset, MMBench-EN. MMBench includes a carefully curated evaluation pipeline with strong quality control and adopts strategies such as CircularEval and LLM-assisted answer matching, making it suitable for robust and fine-grained comparison across multimodal models.

\noindent\textbf{MM-Vet~\cite{mmvet}.} MM-Vet is an open-ended benchmark for evaluating large multimodal models on complicated multimodal tasks that require the integration of multiple abilities. It defines six core vision-language capabilities, including recognition, OCR, knowledge, language generation, spatial awareness, and math, and evaluates 16 capability combinations derived from them. MM-Vet also employs an LLM-based evaluator to score open-ended outputs under a unified metric, making it especially suitable for assessing integrated multimodal competence.

\subsection{Video Benchmarks}

\noindent\textbf{MVBench~\cite{mvbench}.} MVBench is a comprehensive benchmark for evaluating multi-modal video understanding. It contains 2,636 carefully curated multiple-choice question-answer pairs spanning 20 challenging video tasks, and covers four major ability dimensions: temporal understanding, spatial understanding, object-centric understanding, and action-centric understanding. It is designed to assess whether models can effectively comprehend dynamic visual content beyond static image perception.

\noindent\textbf{LongVideo / LongVideoBench~\cite{longvideobench}.} LongVideoBench is a benchmark designed for long-context video understanding. It contains 1,182 long videos with an average duration of approximately 39 minutes and 3,763 manually annotated multiple-choice question-answer pairs. The benchmark evaluates a model's ability to process and reason over long videos, often together with subtitles, making it particularly suitable for testing long-range temporal modeling and long-context multimodal comprehension.

\noindent\textbf{NextQA~\cite{nextqa}.} NextQA is a benchmark for causal and temporal video question answering. Built upon 8,134 videos and more than 52,000 question-answer pairs, it emphasizes higher-order reasoning over videos, including questions about causality, temporal relations, and descriptive understanding. Unlike conventional video QA datasets that focus mainly on short-term recognition, NextQA is designed to evaluate whether models can perform deeper reasoning about events and their dynamics.

\noindent\textbf{VideoMME~\cite{videomme}.} VideoMME is a large-scale benchmark for comprehensive evaluation of video understanding in multimodal large language models. It includes 900 videos and 2,700 multiple-choice questions, with video durations ranging from short to very long clips. The benchmark covers a wide variety of video content and task settings, and is specifically designed to measure a model's ability to understand, reason about, and answer questions over videos of diverse temporal lengths.

\section{Details of Comparison Methods}
\label{appendix:comparison}
\noindent\textbf{FastV~\cite{fastv}.} FastV prunes visual tokens from a selected LLM layer onward by ranking them according to the average attention they receive from textual tokens and discarding the low-ranked ones, thereby reducing the computation of subsequent LLM layers.

\noindent\textbf{LLaVA-PruMerge~\cite{prumerge}.} LLaVA-PruMerge first selects important visual tokens according to the sparsity of the attention between the [CLS] token and patch tokens in the visual encoder, and then merges pruned tokens into the retained ones based on key similarity to preserve visual information.

\noindent\textbf{FasterVLM~\cite{fastervlm}.} FasterVLM ranks visual tokens using the attention from the [CLS] token in the visual encoder, instead of using text-visual attention inside the LLM. It then prunes low-ranked visual tokens before they are fed into the LLM, thereby reducing the computation of the subsequent language model more efficiently.

\noindent\textbf{TRIM~\cite{trim}.} TRIM measures the importance of each image token using the similarity between the text representation and image patch representations in CLIP, adaptively retains high-similarity tokens via an IQR-based selection rule, and appends an aggregated token summarizing the discarded tokens to alleviate information loss.

\noindent\textbf{VisPruner~\cite{vispruner}.} VisPruner first selects important tokens according to visual attention from the visual encoder, and then retains an additional set of diverse tokens by removing redundant ones based on token similarity, so as to better preserve both salient and complementary visual information before feeding the tokens into the LLM. 

\noindent\textbf{SparseVLM~\cite{sparsevlm}.} SparseVLM uses selected text tokens as raters to assess the significance of visual tokens via decoder self-attention, adaptively prunes low-significance tokens based on a rank-based sparsification strategy, and further recycles pruned tokens into compact representations to reduce information loss.

\noindent\textbf{HiRED~\cite{hired}.} HiRED first allocates different token budgets among the image sub-regions according to early-layer CLS attention in the vision encoder, and then retains the most informative tokens within each region based on final-layer CLS attention before feeding them into the LLM. 

\noindent\textbf{DART~\cite{dart}.} DART prunes tokens based on duplication rather than attention-based importance. It first selects a small set of pivot tokens, then measures the cosine similarity between pivot tokens and the remaining visual tokens, and finally retains the tokens with the lowest duplication to the pivots so as to reduce redundancy while preserving visual information.

\noindent\textbf{VisionZIP~\cite{visionzip}.} VisionZIP first selects dominant visual tokens according to attention scores within the vision encoder, and then merges the remaining tokens based on similarity to form contextual tokens, thereby reducing visual redundancy before feeding the tokens into the LLM.

\noindent\textbf{DivPrune~\cite{divprune}.} DivPrune formulates token selection as a max-min diversity problem, aiming to retain a subset of visual tokens with the largest diversity. It selects tokens by maximizing the minimum pairwise distance among them, thereby reducing redundancy and preserving a more representative subset of the original visual tokens. 

\noindent\textbf{PDrop~\cite{pdrop}.} PDrop is a progressive visual token reduction method that divides the LLM into multiple stages and drops a portion of image tokens at the end of each stage according to a predefined ratio, where token selection is based on a lightweight attention-based ranking so that visual redundancy is reduced gradually as the layers deepen. 

\noindent\textbf{SCOPE~\cite{scope}.} SCOPE aims to preserve both salient and semantically complementary visual information. It first models the representational contribution of a selected token set based on feature similarities, then estimates the marginal gain brought by each unselected token. By combining this gain with a saliency score, SCOPE iteratively selects tokens that are both informative and complementary to the current subset, thereby alleviating the semantic incompleteness caused by purely saliency-based pruning.

\noindent\textbf{AdaptPrune~\cite{adaprune}.} AdaptPrune combines attention-based ranking with similarity-based redundancy removal. It analyzes image complexity using effective rank and attention entropy, then adjusts the similarity threshold accordingly so that simple images favor more concentrated high-attention tokens, while complex images retain more diverse tokens. 

\noindent\textbf{PruneSID~\cite{prunesid}.} PruneSID jointly considers token importance and diversity through a two-stage pipeline. It first groups tokens into semantically coherent clusters via principal semantic component analysis, and then removes redundant tokens within each group using non-maximum suppression, with an additional dynamic compression mechanism that adjusts the token budget according to image complexity. 

\noindent\textbf{LearnPruner~\cite{learnpruner}.} LearnPruner first removes visually redundant tokens after the vision encoder using a learnable pruning module, supplemented with a small set of diversity tokens, and then performs query-aware pruning in the middle layer of the LLM based on text-to-vision attention to further discard text-irrelevant tokens. 

\noindent\textbf{AgilePruner~\cite{agilepruner}.} AgilePruner combines attention-based ranking with similarity-based redundancy removal. It analyzes image complexity using effective rank and attention entropy, and then adjusts the pruning threshold accordingly so that simple images retain more concentrated high-attention tokens, while complex images preserve a more diverse token set.

\section{Case Study}
\label{appendix:case_study}

\begin{figure}[h]
    \centering
    \includegraphics[width=\columnwidth]{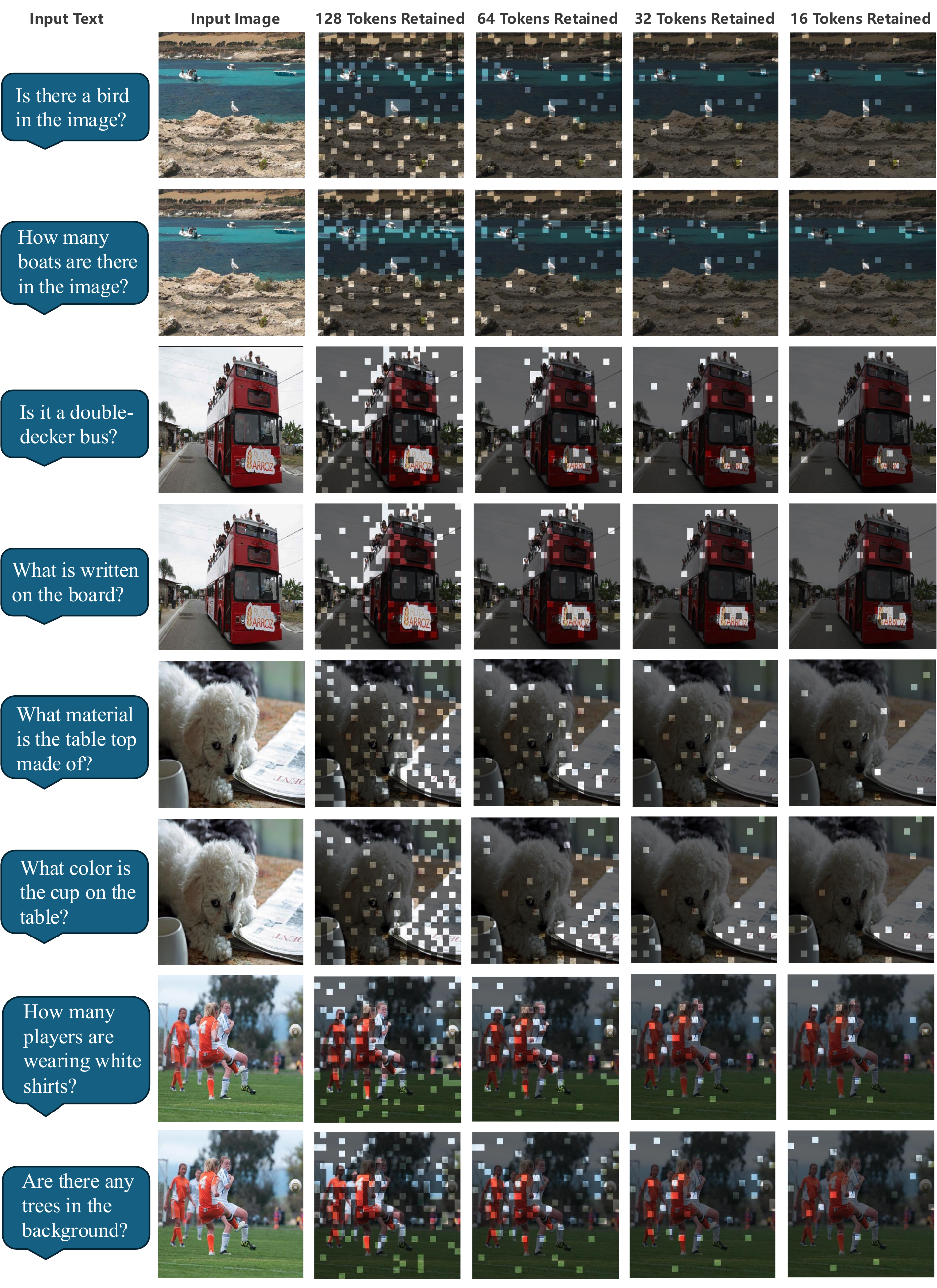}
    \caption{Visualization of ACPruner's pruning results on LLaVA-1.5-7B under different token budgets. The masked patches correspond to visual tokens pruned by ACPruner. For each image, we provide two different text queries to demonstrate the query-relevant nature of ACPruner.}
    \label{fig:case}
\end{figure}

Fig.~\ref{fig:case} visualizes the token pruning results of ACPruner on LLaVA-1.5-7B under different token budgets. For each example, we show the input question, the original image, and the token pruning masks when the budget is set to 128, 64, 32, and 16 tokens. The masked patches indicate the pruned visual tokens.

Several observations can be made from these examples. First, the pruning behavior of ACPruner is clearly query-relevant: for the same image, different questions lead to different retention patterns, even though ACPruner operates entirely in a pre-LLM manner without relying on LLM-side attention. Second, when the token budget is relatively sufficient, ACPruner does not simply preserve a dense cluster of highly similar tokens around a few salient regions. Instead, while retaining the most question-relevant and visually-salient content, it also maintains broader coverage over other informative regions, yielding a compact yet still well-distributed representation of the image. Third, when the token budget becomes extremely limited, ACPruner further concentrates the retained tokens on the most critical visual evidence for the current query, demonstrating a smooth transition from broader semantic coverage to highly selective preservation under aggressive pruning.

Overall, these visualizations provide intuitive support for the design of ACPruner. Rather than merely retaining the individually most salient tokens or uniformly distributing the budget across the image, ACPruner preserves a compact subset of tokens that is both important and collectively informative.

\clearpage
\section*{NeurIPS Paper Checklist}

\begin{enumerate}

\item {\bf Claims}
    \item[] Question: Do the main claims made in the abstract and introduction accurately reflect the paper's contributions and scope?
    \item[] Answer: \answerYes{} 
    \item[] Justification: The abstract and introduction clearly summarize the paper’s contributions, including the formulation of visual token pruning as biased coverage maximization, the design of ACPruner, and its empirical validation across multiple LVLM backbones and task categories, accurately reflecting the scope and claims of the work.
    \item[] Guidelines:
    \begin{itemize}
        \item The answer \answerNA{} means that the abstract and introduction do not include the claims made in the paper.
        \item The abstract and/or introduction should clearly state the claims made, including the contributions made in the paper and important assumptions and limitations. A \answerNo{} or \answerNA{} answer to this question will not be perceived well by the reviewers. 
        \item The claims made should match theoretical and experimental results, and reflect how much the results can be expected to generalize to other settings. 
        \item It is fine to include aspirational goals as motivation as long as it is clear that these goals are not attained by the paper. 
    \end{itemize}

\item {\bf Limitations}
    \item[] Question: Does the paper discuss the limitations of the work performed by the authors?
    \item[] Answer: \answerYes{} 
    \item[] Justification:  We explicitly discuss the limitation in Sec.~\ref{conclusion}.
    \item[] Guidelines:
    \begin{itemize}
        \item The answer \answerNA{} means that the paper has no limitation while the answer \answerNo{} means that the paper has limitations, but those are not discussed in the paper. 
        \item The authors are encouraged to create a separate ``Limitations'' section in their paper.
        \item The paper should point out any strong assumptions and how robust the results are to violations of these assumptions (e.g., independence assumptions, noiseless settings, model well-specification, asymptotic approximations only holding locally). The authors should reflect on how these assumptions might be violated in practice and what the implications would be.
        \item The authors should reflect on the scope of the claims made, e.g., if the approach was only tested on a few datasets or with a few runs. In general, empirical results often depend on implicit assumptions, which should be articulated.
        \item The authors should reflect on the factors that influence the performance of the approach. For example, a facial recognition algorithm may perform poorly when image resolution is low or images are taken in low lighting. Or a speech-to-text system might not be used reliably to provide closed captions for online lectures because it fails to handle technical jargon.
        \item The authors should discuss the computational efficiency of the proposed algorithms and how they scale with dataset size.
        \item If applicable, the authors should discuss possible limitations of their approach to address problems of privacy and fairness.
        \item While the authors might fear that complete honesty about limitations might be used by reviewers as grounds for rejection, a worse outcome might be that reviewers discover limitations that aren't acknowledged in the paper. The authors should use their best judgment and recognize that individual actions in favor of transparency play an important role in developing norms that preserve the integrity of the community. Reviewers will be specifically instructed to not penalize honesty concerning limitations.
    \end{itemize}

\item {\bf Theory assumptions and proofs}
    \item[] Question: For each theoretical result, does the paper provide the full set of assumptions and a complete (and correct) proof?
    \item[] Answer: \answerYes{} 
    \item[] Justification: All theoretical results are formally stated with clear assumptions and complete proofs in Appendix~\ref{appendix:theory} and Appendix~\ref{appendix:nphard}.
    \item[] Guidelines: 
    \begin{itemize}
        \item The answer \answerNA{} means that the paper does not include theoretical results. 
        \item All the theorems, formulas, and proofs in the paper should be numbered and cross-referenced.
        \item All assumptions should be clearly stated or referenced in the statement of any theorems.
        \item The proofs can either appear in the main paper or the supplemental material, but if they appear in the supplemental material, the authors are encouraged to provide a short proof sketch to provide intuition. 
        \item Inversely, any informal proof provided in the core of the paper should be complemented by formal proofs provided in appendix or supplemental material.
        \item Theorems and Lemmas that the proof relies upon should be properly referenced. 
    \end{itemize}

    \item {\bf Experimental result reproducibility}
    \item[] Question: Does the paper fully disclose all the information needed to reproduce the main experimental results of the paper to the extent that it affects the main claims and/or conclusions of the paper (regardless of whether the code and data are provided or not)?
    \item[] Answer: \answerYes{} 
    \item[] Justification: We provide a detailed pseudocode of the proposed method in Appendix~\ref{appendix:pseudocode}. We also provide the detailed hyperparameter settings for reproducing the results in Sec.~\ref{experimental-setup}.
    \item[] Guidelines:
    \begin{itemize}
        \item The answer \answerNA{} means that the paper does not include experiments.
        \item If the paper includes experiments, a \answerNo{} answer to this question will not be perceived well by the reviewers: Making the paper reproducible is important, regardless of whether the code and data are provided or not.
        \item If the contribution is a dataset and\slash or model, the authors should describe the steps taken to make their results reproducible or verifiable. 
        \item Depending on the contribution, reproducibility can be accomplished in various ways. For example, if the contribution is a novel architecture, describing the architecture fully might suffice, or if the contribution is a specific model and empirical evaluation, it may be necessary to either make it possible for others to replicate the model with the same dataset, or provide access to the model. In general. releasing code and data is often one good way to accomplish this, but reproducibility can also be provided via detailed instructions for how to replicate the results, access to a hosted model (e.g., in the case of a large language model), releasing of a model checkpoint, or other means that are appropriate to the research performed.
        \item While NeurIPS does not require releasing code, the conference does require all submissions to provide some reasonable avenue for reproducibility, which may depend on the nature of the contribution. For example
        \begin{enumerate}
            \item If the contribution is primarily a new algorithm, the paper should make it clear how to reproduce that algorithm.
            \item If the contribution is primarily a new model architecture, the paper should describe the architecture clearly and fully.
            \item If the contribution is a new model (e.g., a large language model), then there should either be a way to access this model for reproducing the results or a way to reproduce the model (e.g., with an open-source dataset or instructions for how to construct the dataset).
            \item We recognize that reproducibility may be tricky in some cases, in which case authors are welcome to describe the particular way they provide for reproducibility. In the case of closed-source models, it may be that access to the model is limited in some way (e.g., to registered users), but it should be possible for other researchers to have some path to reproducing or verifying the results.
        \end{enumerate}
    \end{itemize}

\item {\bf Open access to data and code}
    \item[] Question: Does the paper provide open access to the data and code, with sufficient instructions to faithfully reproduce the main experimental results, as described in supplemental material?
    \item[] Answer: \answerNo{} 
    \item[] Justification: We will release the code once the paper is accepted.
    \item[] Guidelines:
    \begin{itemize}
        \item The answer \answerNA{} means that paper does not include experiments requiring code.
        \item Please see the NeurIPS code and data submission guidelines (\url{https://neurips.cc/public/guides/CodeSubmissionPolicy}) for more details.
        \item While we encourage the release of code and data, we understand that this might not be possible, so \answerNo{} is an acceptable answer. Papers cannot be rejected simply for not including code, unless this is central to the contribution (e.g., for a new open-source benchmark).
        \item The instructions should contain the exact command and environment needed to run to reproduce the results. See the NeurIPS code and data submission guidelines (\url{https://neurips.cc/public/guides/CodeSubmissionPolicy}) for more details.
        \item The authors should provide instructions on data access and preparation, including how to access the raw data, preprocessed data, intermediate data, and generated data, etc.
        \item The authors should provide scripts to reproduce all experimental results for the new proposed method and baselines. If only a subset of experiments are reproducible, they should state which ones are omitted from the script and why.
        \item At submission time, to preserve anonymity, the authors should release anonymized versions (if applicable).
        \item Providing as much information as possible in supplemental material (appended to the paper) is recommended, but including URLs to data and code is permitted.
    \end{itemize}

\item {\bf Experimental setting/details}
    \item[] Question: Does the paper specify all the training and test details (e.g., data splits, hyperparameters, how they were chosen, type of optimizer) necessary to understand the results?
    \item[] Answer: \answerYes{} 
    \item[] Justification: We specify all relevant experimental details, including datasets, backbone models, and hyperparameter choices in Sec.~\ref{experimental-setup}. 
    \item[] Guidelines:
    \begin{itemize}
        \item The answer \answerNA{} means that the paper does not include experiments.
        \item The experimental setting should be presented in the core of the paper to a level of detail that is necessary to appreciate the results and make sense of them.
        \item The full details can be provided either with the code, in appendix, or as supplemental material.
    \end{itemize}

\item {\bf Experiment statistical significance}
    \item[] Question: Does the paper report error bars suitably and correctly defined or other appropriate information about the statistical significance of the experiments?
    \item[] Answer: \answerNo{} 
    \item[] Justification: Following prior studies, all experiments use deterministic decoding with temperature set to 0. As a result, there is no source of randomness in the evaluation, and error bars or statistical significance measures are not applicable.
    \item[] Guidelines:
    \begin{itemize}
        \item The answer \answerNA{} means that the paper does not include experiments.
        \item The authors should answer \answerYes{} if the results are accompanied by error bars, confidence intervals, or statistical significance tests, at least for the experiments that support the main claims of the paper.
        \item The factors of variability that the error bars are capturing should be clearly stated (for example, train/test split, initialization, random drawing of some parameter, or overall run with given experimental conditions).
        \item The method for calculating the error bars should be explained (closed form formula, call to a library function, bootstrap, etc.)
        \item The assumptions made should be given (e.g., Normally distributed errors).
        \item It should be clear whether the error bar is the standard deviation or the standard error of the mean.
        \item It is OK to report 1-sigma error bars, but one should state it. The authors should preferably report a 2-sigma error bar than state that they have a 96\% CI, if the hypothesis of Normality of errors is not verified.
        \item For asymmetric distributions, the authors should be careful not to show in tables or figures symmetric error bars that would yield results that are out of range (e.g., negative error rates).
        \item If error bars are reported in tables or plots, the authors should explain in the text how they were calculated and reference the corresponding figures or tables in the text.
    \end{itemize}

\item {\bf Experiments compute resources}
    \item[] Question: For each experiment, does the paper provide sufficient information on the computer resources (type of compute workers, memory, time of execution) needed to reproduce the experiments?
    \item[] Answer: \answerYes{} 
    \item[] Justification: We indicate the computation resources in Sec.~\ref{experimental-setup} and provide practical runtime analysis in Sec.~\ref{efficiency}.
    \item[] Guidelines:
    \begin{itemize}
        \item The answer \answerNA{} means that the paper does not include experiments.
        \item The paper should indicate the type of compute workers CPU or GPU, internal cluster, or cloud provider, including relevant memory and storage.
        \item The paper should provide the amount of compute required for each of the individual experimental runs as well as estimate the total compute. 
        \item The paper should disclose whether the full research project required more compute than the experiments reported in the paper (e.g., preliminary or failed experiments that didn't make it into the paper). 
    \end{itemize}
    
\item {\bf Code of ethics}
    \item[] Question: Does the research conducted in the paper conform, in every respect, with the NeurIPS Code of Ethics \url{https://neurips.cc/public/EthicsGuidelines}?
    \item[] Answer: \answerYes{} 
    \item[] Justification: This work fully complies with the NeurIPS Code of Ethics.
    \item[] Guidelines:
    \begin{itemize}
        \item The answer \answerNA{} means that the authors have not reviewed the NeurIPS Code of Ethics.
        \item If the authors answer \answerNo, they should explain the special circumstances that require a deviation from the Code of Ethics.
        \item The authors should make sure to preserve anonymity (e.g., if there is a special consideration due to laws or regulations in their jurisdiction).
    \end{itemize}

\item {\bf Broader impacts}
    \item[] Question: Does the paper discuss both potential positive societal impacts and negative societal impacts of the work performed?
    \item[] Answer: \answerNA{} 
    \item[] Justification: This work focuses on algorithmic improvements for efficient visual token pruning in LVLMs and does not involve deployment in sensitive contexts or affect human subjects directly. Consequently, discussion of broader positive or negative societal impacts is not applicable.
    \item[] Guidelines:
    \begin{itemize}
        \item The answer \answerNA{} means that there is no societal impact of the work performed.
        \item If the authors answer \answerNA{} or \answerNo, they should explain why their work has no societal impact or why the paper does not address societal impact.
        \item Examples of negative societal impacts include potential malicious or unintended uses (e.g., disinformation, generating fake profiles, surveillance), fairness considerations (e.g., deployment of technologies that could make decisions that unfairly impact specific groups), privacy considerations, and security considerations.
        \item The conference expects that many papers will be foundational research and not tied to particular applications, let alone deployments. However, if there is a direct path to any negative applications, the authors should point it out. For example, it is legitimate to point out that an improvement in the quality of generative models could be used to generate Deepfakes for disinformation. On the other hand, it is not needed to point out that a generic algorithm for optimizing neural networks could enable people to train models that generate Deepfakes faster.
        \item The authors should consider possible harms that could arise when the technology is being used as intended and functioning correctly, harms that could arise when the technology is being used as intended but gives incorrect results, and harms following from (intentional or unintentional) misuse of the technology.
        \item If there are negative societal impacts, the authors could also discuss possible mitigation strategies (e.g., gated release of models, providing defenses in addition to attacks, mechanisms for monitoring misuse, mechanisms to monitor how a system learns from feedback over time, improving the efficiency and accessibility of ML).
    \end{itemize}
    
\item {\bf Safeguards}
    \item[] Question: Does the paper describe safeguards that have been put in place for responsible release of data or models that have a high risk for misuse (e.g., pre-trained language models, image generators, or scraped datasets)?
    \item[] Answer: \answerNA{} 
    \item[] Justification: This work does not release any models or datasets that pose a high risk of misuse. The paper focuses solely on algorithmic improvements for visual token pruning in LVLMs, and therefore no specific safeguards are required.
    \item[] Guidelines:
    \begin{itemize}
        \item The answer \answerNA{} means that the paper poses no such risks.
        \item Released models that have a high risk for misuse or dual-use should be released with necessary safeguards to allow for controlled use of the model, for example by requiring that users adhere to usage guidelines or restrictions to access the model or implementing safety filters. 
        \item Datasets that have been scraped from the Internet could pose safety risks. The authors should describe how they avoided releasing unsafe images.
        \item We recognize that providing effective safeguards is challenging, and many papers do not require this, but we encourage authors to take this into account and make a best faith effort.
    \end{itemize}

\item {\bf Licenses for existing assets}
    \item[] Question: Are the creators or original owners of assets (e.g., code, data, models), used in the paper, properly credited and are the license and terms of use explicitly mentioned and properly respected?
    \item[] Answer: \answerYes{} 
    \item[] Justification: All datasets and models used in this work are publicly available and properly cited. 
    \item[] Guidelines:
    \begin{itemize}
        \item The answer \answerNA{} means that the paper does not use existing assets.
        \item The authors should cite the original paper that produced the code package or dataset.
        \item The authors should state which version of the asset is used and, if possible, include a URL.
        \item The name of the license (e.g., CC-BY 4.0) should be included for each asset.
        \item For scraped data from a particular source (e.g., website), the copyright and terms of service of that source should be provided.
        \item If assets are released, the license, copyright information, and terms of use in the package should be provided. For popular datasets, \url{paperswithcode.com/datasets} has curated licenses for some datasets. Their licensing guide can help determine the license of a dataset.
        \item For existing datasets that are re-packaged, both the original license and the license of the derived asset (if it has changed) should be provided.
        \item If this information is not available online, the authors are encouraged to reach out to the asset's creators.
    \end{itemize}

\item {\bf New assets}
    \item[] Question: Are new assets introduced in the paper well documented and is the documentation provided alongside the assets?
    \item[] Answer: \answerNA{} 
    \item[] Justification: We do not release new assets at this stage. 
    \item[] Guidelines: 
    \begin{itemize}
        \item The answer \answerNA{} means that the paper does not release new assets.
        \item Researchers should communicate the details of the dataset\slash code\slash model as part of their submissions via structured templates. This includes details about training, license, limitations, etc. 
        \item The paper should discuss whether and how consent was obtained from people whose asset is used.
        \item At submission time, remember to anonymize your assets (if applicable). You can either create an anonymized URL or include an anonymized zip file.
    \end{itemize}

\item {\bf Crowdsourcing and research with human subjects}
    \item[] Question: For crowdsourcing experiments and research with human subjects, does the paper include the full text of instructions given to participants and screenshots, if applicable, as well as details about compensation (if any)? 
    \item[] Answer: \answerNA{} 
    \item[] Justification: This work does not involve crowdsourcing nor research with human subjects.
    \item[] Guidelines:
    \begin{itemize}
        \item The answer \answerNA{} means that the paper does not involve crowdsourcing nor research with human subjects.
        \item Including this information in the supplemental material is fine, but if the main contribution of the paper involves human subjects, then as much detail as possible should be included in the main paper. 
        \item According to the NeurIPS Code of Ethics, workers involved in data collection, curation, or other labor should be paid at least the minimum wage in the country of the data collector. 
    \end{itemize}

\item {\bf Institutional review board (IRB) approvals or equivalent for research with human subjects}
    \item[] Question: Does the paper describe potential risks incurred by study participants, whether such risks were disclosed to the subjects, and whether Institutional Review Board (IRB) approvals (or an equivalent approval/review based on the requirements of your country or institution) were obtained?
    \item[] Answer: \answerNA{} 
    \item[] Justification: This work does not involve crowdsourcing nor research with human subjects.
    \item[] Guidelines:
    \begin{itemize}
        \item The answer \answerNA{} means that the paper does not involve crowdsourcing nor research with human subjects.
        \item Depending on the country in which research is conducted, IRB approval (or equivalent) may be required for any human subjects research. If you obtained IRB approval, you should clearly state this in the paper. 
        \item We recognize that the procedures for this may vary significantly between institutions and locations, and we expect authors to adhere to the NeurIPS Code of Ethics and the guidelines for their institution. 
        \item For initial submissions, do not include any information that would break anonymity (if applicable), such as the institution conducting the review.
    \end{itemize}

\item {\bf Declaration of LLM usage}
    \item[] Question: Does the paper describe the usage of LLMs if it is an important, original, or non-standard component of the core methods in this research? Note that if the LLM is used only for writing, editing, or formatting purposes and does \emph{not} impact the core methodology, scientific rigor, or originality of the research, declaration is not required.
    \item[] Answer: \answerNA{} 
    \item[] Justification: The core method development in this research does not involve LLMs as any important, original, or non-standard components.
    \item[] Guidelines:
    \begin{itemize}
        \item The answer \answerNA{} means that the core method development in this research does not involve LLMs as any important, original, or non-standard components.
        \item Please refer to our LLM policy in the NeurIPS handbook for what should or should not be described.
    \end{itemize}

\end{enumerate}

\end{document}